# AI for Water Sustainability: Global Water Quality Assessment and Prediction with Explainable AI with LLM Chatbot for Insights

Biplov Paneru,

Dept. of Electronics and communication Engineering,

Nepal Engineering College, Pokhara University, Nepal

Bishwash Paneru,

Dept.of applied science and chemical Engineering,

IOE Pulchowk Campus, Tribhuvan University, Nepal

***Abstract****:* Ensuring safe water supplies requires effective water quality monitoring, especially in developing countries like Nepal, where contamination risks are high. This paper introduces various hybrid deep learning models to predict on the CCME dataset with multiple water quality parameters from Canada, China, the UK, the USA, and Ireland, with 2.82 million data records feature-engineered and evaluated using them. Models such as CatBoost, XGBoost, and Extra Trees, along with neural networks combining CNN and LSTM layers, are used to capture temporal and spatial patterns in the data. The model demonstrated notable accuracy improvements, aiding proactive water quality control. CatBoost, XGBoost, and Extra Trees Regressor predicted Water Quality Index (WQI) values with an average RMSE of 1.2 and an $R^2$ score of 0.99. Additionally, classifiers achieved 99% accuracy, cross-validated across models. SHAP analysis showed the importance of indicators like F.R.C. and orthophosphate levels in hybrid architectures' classification decisions. The practical application is demonstrated along with a chatbot application for water quality insights.

## 1. Introduction

Due to the serious health and environmental risks posed by water contamination, an immediate assessment and remediation are required. Regular monitoring of essential water quality parameters, such as pH, dissolved oxygen, temperature, turbidity, etc., provide crucial information on pollution levels and the general health of rivers [1]. Additionally, in order to lessen water pollution, regular maintenance should be done on ponds and other water sources. Potable water must be devoid of harmful organisms and have minimal amounts of substances that are very poisonous to humans or animals. Along with ponds and rivers, the responsible organizations should create a plan to protect and lessen water pollution in various water sources [9]. Monitoring water quality is vital to ensuring a clean and safe water supply, especially in developing countries like Nepal, where water sources are susceptible to pollution. Effective water quality management is essential to address this challenge, and it relies on accurate and timely predictions of water quality indices (WQI). Water is one of the most essential natural resources for all life on Earth. Having access to clean water is a basic human requirement. The past few decades have seen a significant drop in water quality due to pollution and a host of other issues [14]. Since most living things get their physical health from the water they drink, maintaining the water's purity and cleanliness is crucial since contaminated water can have harmful effects on human health as well as the environment. This makes measuring, regulating, and keeping an eye on the water quality absolutely necessary [15].

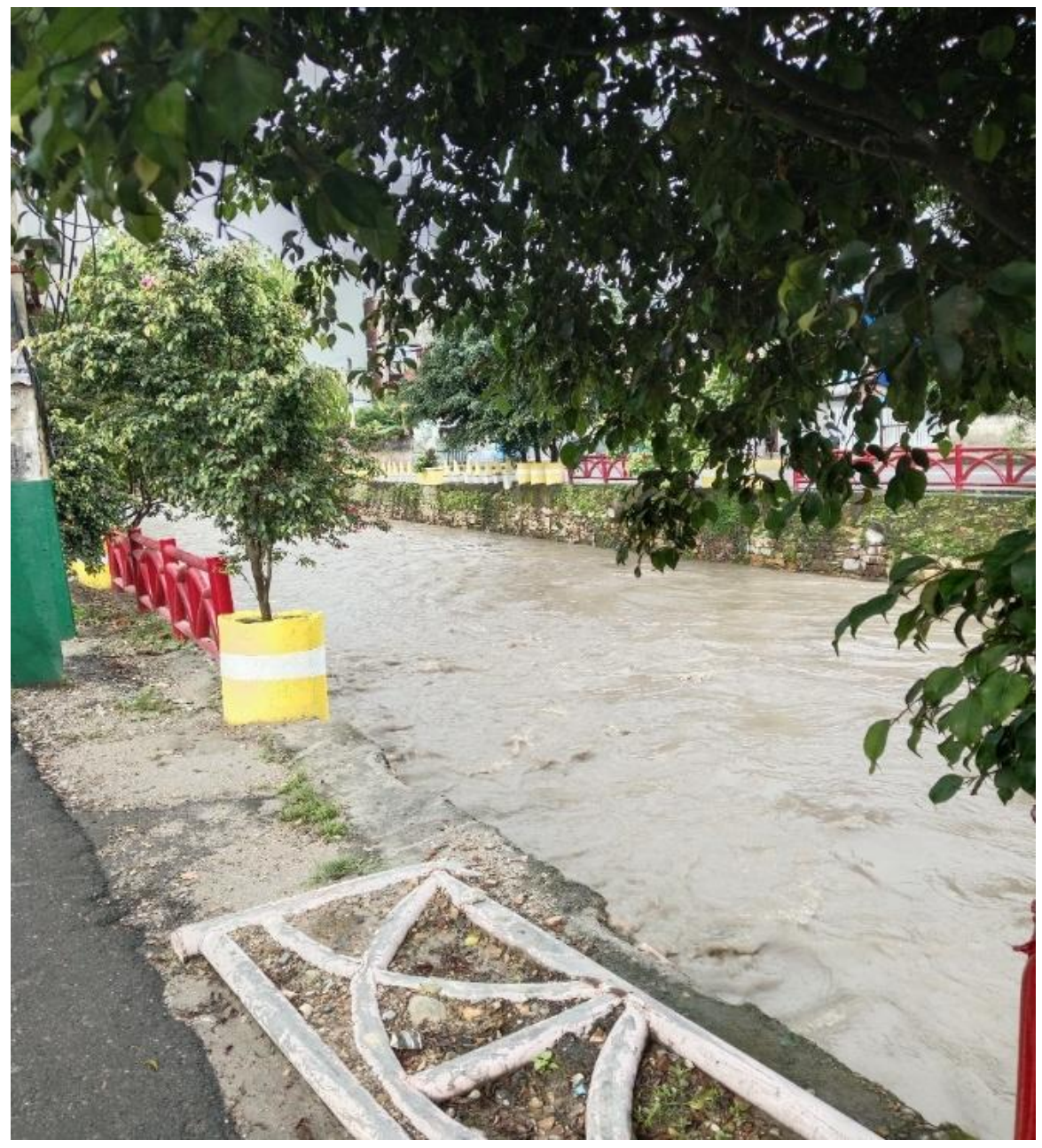

Fig 1. Flooded condition in Kathmandu, Nepal

The world is slowly approaching a water shortage and massive dryness as illustrated in figure 1. Such situation isn’t far away. Traditional methods often fall short in capturing the complex temporal and spatial patterns inherent in water quality data. With more than 80% of wastewater being dumped into natural water systems worldwide, water pollution is a serious problem. In underdeveloped nations, where it is believed that over 95% of wastewater has been dumped into water bodies without treatment, the situation is more worse (UN, 2017). Such contamination restricts the appropriate usage of water resources and poses a serious risk to human and environmental health [11]. A grave concern that humanity faces is the declining quality of natural water resources such as lakes, streams, and estuaries. Water contamination has far-reaching consequences that affect all facets of existence. As a result, effective management of water resources is essential to maximizing water quality [13].

In order to forecast the seasonal water quality in Nepal, this study proposes a hybrid deep learning model that makes use of a sizable dataset of water quality variables. Utilizing both temporal and spatial patterns in the input, the model combines recurrent neural networks (RNN) with convolutional neural networks (CNN). The outcomes show a considerable increase in forecast accuracy over traditional methods, offering a trustworthy instrument for preventive water quality management.

## 1.1 Literature Review

In recent years, Machine Learning (ML) and Artificial Intelligence (AI) have shown increasing applicability in water quality prediction, classification, and monitoring systems. Diverse models ranging from traditional ML algorithms to deep learning and explainable AI techniques have been used across multiple geographical and hydrological contexts. Below, we categorize and summarize key related works.

1. Time Series Forecasting and Deep Learning for Water Quality

Studies in this category focus on using temporal data with models like ARIMA, LSTM, and other RNN variants:

- Bashyal et al. (2023) applied LSTM with a sequence-to-sequence structure and ARIMA to predict Bagmati River water quality. Additionally, a DNN was used for location-based prediction using latitude and longitude inputs [1].
- Del Castillo et al. (2024) used time series analysis (TSA) with ML models like ANN, SVM, and ANFIS to cluster monitoring stations based on long-term water quality data from the Santiago River, Mexico. Optimal monitoring stations were identified via BTS selection [17].
- Rammohan et al. (2024) forecasted groundwater quality in West Texas by analyzing TDS levels using time-series data from the TWDB, achieving reliable long-term predictions [23].
- Choden et al. (2022) explored water quality prediction in Bhutan’s Wangchu River using MLR, ANN, and ANFIS. ANN using Bayesian Regularization (trainBR) performed best with CC = 99.8% [19].

## 2. Supervised Machine Learning for WQI Prediction and Classification

These works implement classic ML classifiers such as Random Forest, SVM, and boosting models:

- Abbas et al. (2024) utilized SVM, RF, Gradient Boosting, KNN, and Decision Trees for WQI prediction in Sindh. RF and GB performed best with 95% and 96% accuracy, respectively [3].
- Sangwan et al. (2024) evaluated eight ML models using accuracy, recall, precision, and other classification metrics based on key physical and chemical parameters [7].
- Ashfaq et al. (2024) applied six ML algorithms for WQC using 16 input parameters. RF outperformed other models [14].
- Venkatesh et al. (2024) implemented Ridge Regression and ensemble regressors to predict groundwater quality. Ridge Regression achieved $R^2 = 0.999$ [8].
- Shams et al. (2023) showed that the GB model and MLP regressor were most effective, achieving up to 99.8% $R^2$ in WQC prediction [6].
- Nasir et al. (2022) employed multiple classifiers including CatBoost and stacking ensembles to classify WQI with 100% accuracy using ensemble meta-classifiers [21].

## 3. Ensemble Learning and Optimization Approaches

These studies explore model enhancement through ensemble techniques, boosting, or hyperparameter optimization:

- Alqahtani et al. (2024) compared RF, ANN, and GEP using metrics like $R^2$ and NSE. Optimized RF showed the highest performance at $R^2 = 0.96$ [4].
- Makumbura et al. (2024) used SHAP analysis with ensemble models (RF, LightGBM, XGBoost) to understand water quality drivers like COD and BOD, achieving $R^2 = 0.992$ [20].

- Malek et al. (2022) enhanced model generalizability using bagging and boosting. Gradient Boosting performed best, identifying TSS as the most influential parameter [18].
- AutoML system study (2024) proposed a stacked ensemble model with SHAP interpretation, achieving 97% accuracy and outperforming seven other ML models [25].

4. Explainable AI (XAI) and Feature Interpretation in Water Prediction

These works focused on using XAI tools like SHAP to provide transparency in model predictions:

- Nallakaruppan et al. (2024) integrated SHAP visualizations (force plot, summary plot, etc.) with RF to explain drinkability classification. The RF model showed an F1-score of 0.9999 [15].
- Makumbura et al. (2024) used SHAP to explain the influence of parameters in ensemble models, identifying DO, BOD, and COD as critical contributors [20].
- Our proposed work leverages LIME-based interpretability for a hybrid deep learning model combining CNN and LSTM to classify water quality into three categories, making real-time monitoring viable.

5. Sensor-Based and Image-Based Monitoring

These works incorporate environmental sensing or image data into water quality models:

- Iffaty et al. (2023) trained a model using 80 water images labeled into turbidity classes. Despite dataset limitations, the system enabled economical real-time turbidity assessment [16].
- Rammohan et al. (2024) combined GIS tools and ML (XGBoost, LSTM) to predict and visualize GQI across Kanchipuram, Tamil Nadu [22].

6. Hybrid and Deep Learning Models

These studies used combinations of neural networks and deep architectures:

- Bashyal et al. (2023) proposed a sequence-to-sequence LSTM with repeat layers for forecasting and DNN for location-based prediction [1].
- Del Castillo et al. (2024) incorporated ANFIS and ANN for spatial-temporal modeling of river data [17].
- Our proposed approach uses a hybrid Conv1D + LSTM model to enhance seasonal WQI prediction and classification, optimized with cross-validation and real-time deployment capabilities.

7. Groundwater Quality Prediction and Spatial Analysis

These studies specifically focus on groundwater datasets and regional modeling:

- Venkatesh et al. (2024) applied ML for groundwater quality monitoring near waste disposal zones, showing that RR provides the best predictions [8].
- Rammohan et al. (2024) examined groundwater across six counties in Texas, using historical TDS levels for prediction [23].
- Rammohan et al. (2024) also evaluated water quality in Kanchipuram using integrated ML with GIS visualizations and verified predictions using new well samples [22].

8. National-Level and Government Dataset Applications

These papers leverage public data for model building and policy-level insights:

- Kuroki et al. (2024) used government-provided tap water and source water data to train SVM, RF, and LightGBM models, targeting E. coli detection [2].

- Khan et al. (2016) evaluated water data from the USGS (NWIS), comparing regression models using MSE and RMSE [13].

9. Ecosystem and Environmental Impact Analysis

Some studies linked water quality degradation to ecological outcomes:

- Bai et al. (2024) used RF to analyze Beibu Gulf's water quality and pollution trends, linking seasonal exceedances to coastal ecosystem strain [24].
- Del Castillo et al. (2024) analyzed the long-term impacts of river pollution on ecosystem services using advanced ML and clustering techniques [17].

### 1.2 Summary and Gap Analysis

Across these studies, Random Forest, Gradient Boosting, and hybrid deep learning models consistently yield high performance in water quality prediction tasks. While many works focus on single-region or parameter-specific models, our proposed research addresses this limitation by combining ensemble modeling, hybrid CNN-LSTM architecture, and explainable AI (LIME) **to** provide seasonal, real-time, and stratified predictions**.** Furthermore, we offer a user-facing interface and RAG-powered chatbot for enhanced stakeholder engagement and data-informed decision making in the context of Kathmandu Valley.

Previous studies on water quality prediction have employed various machine learning approaches, including support vector machines (SVM), decision trees, and neural networks. These models typically focus on specific parameters or regions, limiting their generalizability. Recent advancements in deep learning have shown promise in capturing intricate patterns in environmental data. CNNs are effective for spatial data analysis, while RNNs excel in handling temporal sequences. A hybrid approach can harness the strengths of both architectures, offering improved predictive performance for water quality monitoring. We aim to introduce various ensemble model-based analyses for seasonal water quality that have variations due to seasons-based acquisition and perform interpretability analysis using LIME explainable AI. We utilize also two deep learning hybrid models in WQI prediction and classification into three sub-categories: average, good, and poor. Based on the data volume obtained from various wells in Nepal. The models are aimed at being stratified, cross-validated, and nested, cross-validated to monitor their performance over various folds and divisions so that they can be fitted for real-time integration with application systems and other edge systems for monitoring water quality. The multi-featured application designed and integrated with the two different models for different purposes (regression and classification) aims to provide users with an extensive service to predict water quality in Kathmandu using new data and help to classify the water into 3 categories. Along with it, WQI-specific values can be predicted, which offers water experts easy analysis on new data. The RAG-implemented chatbot showcases the potential for gaining insights and knowledge on model predictions about WQI values with enhanced information and guidance for water quality and sustainability practices.

### 1.3 Significance of research (Global Water Quality Scenario)

In the Kathmandu Valley, the Bagmati River and its six tributaries (Bishnumati, Manohara, Dhobikhola, Nakhhu, Balkhu, and Tukucha) are severely polluted. This is due to urbanization, industrial waste, and sewage inflows. The water is visibly black and foul-smelling in urban stretches, severely impacting both aquatic ecosystems and public health. Various stakeholders are attempting restoration through infrastructure projects like sewage diversion and riverbank stabilization, but pollution continues to rise [26].

Globally, water pollution remains a serious crisis:

- 2 million tons of waste are discharged into water bodies daily.
- 2.5 billion people have been lacking adequate sanitation.
- 1.2 billion practice open defecation, leading to widespread contamination.
- Waterborne diseases are the leading cause of death in children under five, with 2.2 million deaths from diarrhea annually.

Water pollution from industry, mining, and agriculture is widespread. Harmful substances like mercury, lead, and cyanide from industrial sources contaminate water. Agricultural runoff, especially nitrates and phosphates, heavily contributes to global water pollution, with increasing nitrate levels reported globally. Infrastructure like dams has also altered natural water flows, reducing sediment transport and impacting ecosystems. While widely referred to as a global water crisis, water challenges are fundamentally local in nature, varying across regions due to differing environmental, social, and governance contexts. Case studies reveal this reality: groundwater depletion in India, salinity in coastal Bangladesh, arsenic contamination in South Asia, water poverty in Sub-Saharan Africa, and sewage pollution in the UK. These issues demand tailored, region-specific solutions, not just broad global policies. Effective water management requires localized action, informed governance, and community-level strategies rather than one-size-fits-all approaches driven by global narratives [27]. Urban water scarcity is a growing global threat, intensified by climate change and rapid urbanization. Currently, around 400 million people in cities face perennial water shortages, and this number is projected to rise to 1 billion by 2050. Unplanned urban growth, such as seen in Cape Town and Mexico City, has worsened water availability, with climate-induced events like floods contaminating groundwater sources. Studies indicate that with a 1.5°C rise in global temperature, 357 million urban residents could face extreme droughts, and with 2°C, the number could nearly double to 696 million. As global temperatures climb, urban centers are heading toward an increasingly dry and water-stressed future [28].

The Middle East and North Africa (MENA) region, inherently arid and water-scarce, is the most water-stressed region in the world. This chronic scarcity is driven by natural climatic conditions, limited freshwater availability, and increasing demand from growing populations and economic activities. Climate change is exacerbating the situation by intensifying droughts, raising temperatures, and altering precipitation patterns, which collectively deepen the gap between water supply and demand [29].

1. Water Scarcity and Stress: Sixteen of the world’s 25 most water-stressed countries are in MENA. Water stress is high due to limited freshwater sources and growing consumption needs.

2. Regional Variation: Countries with access to renewable surface water (e.g., Egypt with the Nile) fare better than others dependent on costly, nonconventional sources like desalination or groundwater.
3. Desalination Dependence: Gulf Cooperation Council (GCC) countries rely heavily on desalination, accounting for 45% of the global desalinated water. While crucial, desalination is energy-intensive, expensive, and may have long-term environmental implications.
4. Groundwater Depletion: Aquifers are overexploited, especially in areas without surface water access. Poor regulation and data collection hinder sustainable groundwater management.
5. Transboundary Water Conflicts: Disputes over major rivers such as the Nile and Tigris-Euphrates are escalating due to dam construction and unilateral water projects, particularly involving Ethiopia, Turkey, and Iran, threatening regional cooperation and stability.
6. Water Quality Concerns: Pollution from sewage, agricultural runoff, and industrial waste affects rivers like the Jordan and Euphrates, leading to public health crises in underserved communities (e.g., cholera in Syria).
7. Climate Change Impacts: Rising temperatures increase evapotranspiration and water demand in agriculture, urban, and industrial sectors. The region has witnessed record-breaking heatwaves (e.g., 152°F in Iran, 2023), damaging crops, raising energy needs, and stressing water systems.
8. Extreme Weather and Infrastructure Risks: Increased frequency of severe weather events—including floods and cyclones—have damaged critical water infrastructure, as seen in Libya's dam collapse during Storm Daniel in 2023.
9. Sea Level Rise: Coastal infrastructure, such as desalination and wastewater treatment plants, faces long-term threats from sea level rise and saltwater intrusion, impacting water supply and aquifer sustainability.
10. Growing Urban Demand: Rapid urbanization and high population growth, especially in countries like Iraq, Sudan, and Palestine, will significantly elevate water demand, straining existing infrastructure.

Water quality is a critical determinant of public health, ecosystem balance, and sustainable development, particularly in water-stressed regions like the Middle East and North Africa (MENA). Traditional methods of water monitoring are often reactive, resource-intensive, and limited in scalability. In this context, predictive modeling using Explainable Artificial Intelligence (XAI), integrated with interactive chatbots, offers a transformative approach for proactive water quality management.XAI-based water quality prediction models leverage deep learning and statistical algorithms to forecast key water parameters—such as pH, turbidity, dissolved oxygen, and total dissolved solids—based on real-time or historical data. These models not only provide accurate predictions but also offer interpretability, enabling stakeholders to understand the impact of individual features (e.g., temperature, rainfall, chemical concentration) on water quality outcomes. This transparency builds trust in AI systems and supports informed decision-making by environmental agencies and policymakers [30].

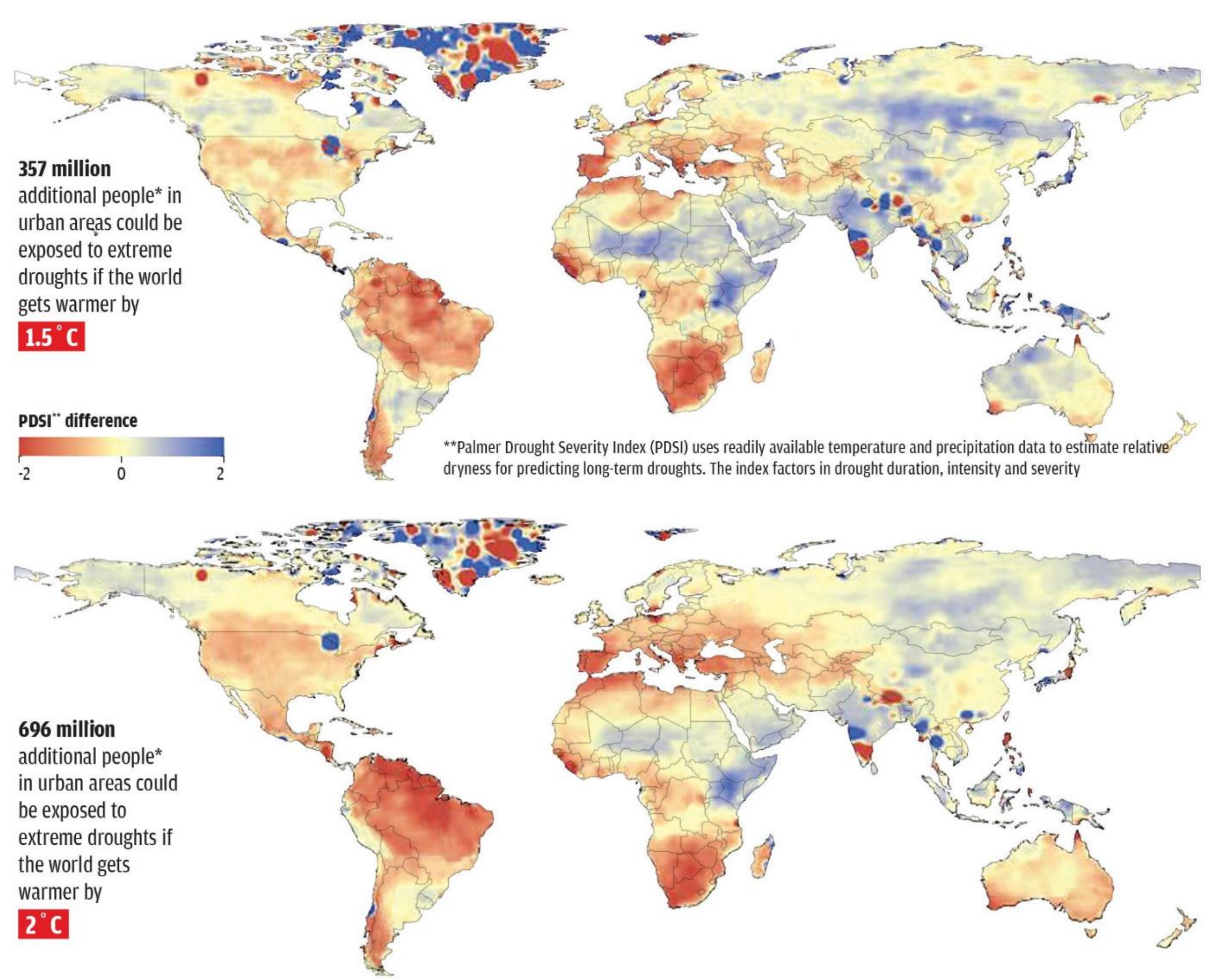

Fig 1(b): Areas to be affected by water scarcity [28]

In parallel, the integration of chatbot systems for insights development acts as an accessible interface between complex AI models and non-technical users. These chatbots, embedded with Natural Language Processing (NLP) capabilities, can interpret user queries, visualize model outputs, explain contributing factors to water pollution, and provide actionable recommendations in real time. This democratizes access to predictive insights and promotes community-level awareness and engagement in water conservation.

## 2. Methodology

### CCMI water quality value:

For calculating water quality of various countries to create a unified water quality measure model for world the data from nature publication [31] was cleaned and combined with various countries data into a unified set. The CCME Water Quality Index (WQI)**,** developed by the Canadian Council of Ministers of the Environment**,** is a widely used composite indicator **to** summarize water quality data into a single number**.** Its value and significance span **both** general water science and machine learning (ML)-based water quality prediction**.** The data repository [31] comprises separate country-specific datasets in CSV format—each containing eight key water quality parameters collected from regional monitoring stations. These parameters include Turbidity, pH, Electrical Conductivity, Total Alkalinity, Total Hardness, Total Iron, Total Ammonia, and Chloride, which are essential indicators for assessing the overall health and safety of water resources. Accompanying these datasets is the Summary_country.xlsx file, which provides descriptive statistics such as the minimum, maximum, mean, and standard deviation for each parameter across the different countries. This summary sheet enables a comparative analysis of water quality trends and variability across regions. The collection is valuable for regional environmental assessment, supports the development of predictive models using Explainable AI (XAI), and is suitable for visualization dashboards or decision-support tools focused on sustainable water management and policy development.

Table 1: CCMI significance In General Water Science Field

| Aspect | Significance |
| --- | --- |
| Standardized Index | CCME-WQI provides a standard method to interpret water quality across regions and parameters. |
| Holistic Measure | Combines multiple parameters (e.g., pH, turbidity, nitrate, DO) into a single composite score. |
| Score Range | Values range from 0 to 100, where:<br>- 95–100 = Excellent<br>- 80–94 = Good<br>- 65–79 = Fair<br>- 45–64 = Marginal<br>- 0–44 = Poor |
| Decision-Making Tool | Used by governments, NGOs, and environmental agencies to monitor and communicate water health. |
| Temporal/Spatial Trends | Enables tracking of changes over time or across geographical locations. |
| Public Communication | Translates complex data into a form that's easily understandable by non-experts. |

**Orthophosphate:**

With only one phosphate ion ($PO_4^{3-}$), orthophosphate is the most basic form of phosphate and is easily absorbed by biological systems. Because of fertilizer runoff, detergents, and sewage discharge, it is frequently found in natural streams and wastewater. Orthophosphate is an essential nutrient for the growth of plants and algae, and too much of it can cause eutrophication, which depletes oxygen in water bodies and results in toxic algal blooms.

## 2.2 Preprocessing and handling data imbalance handling with downsampling

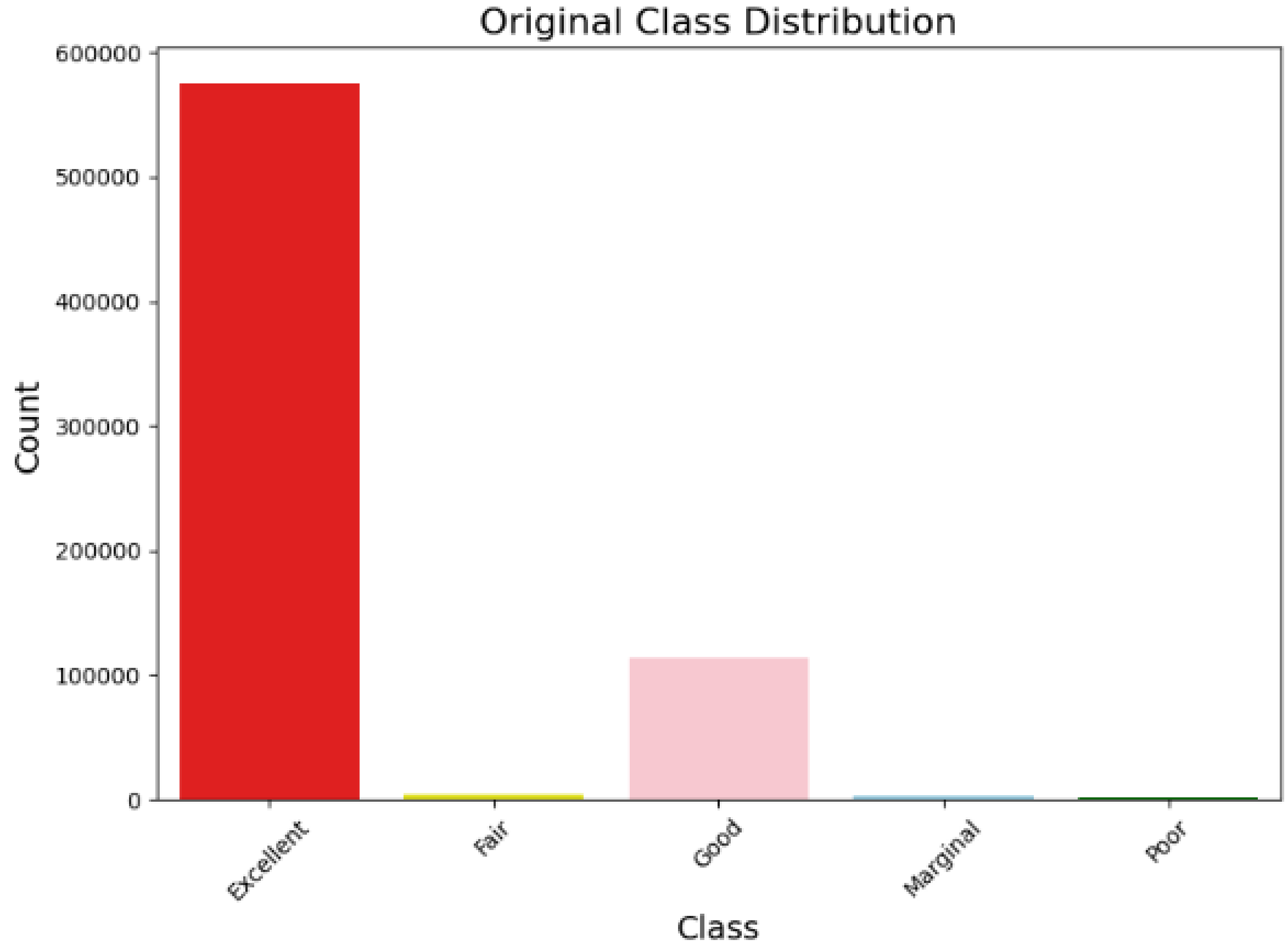

**Fig 2:** Classes original distribution

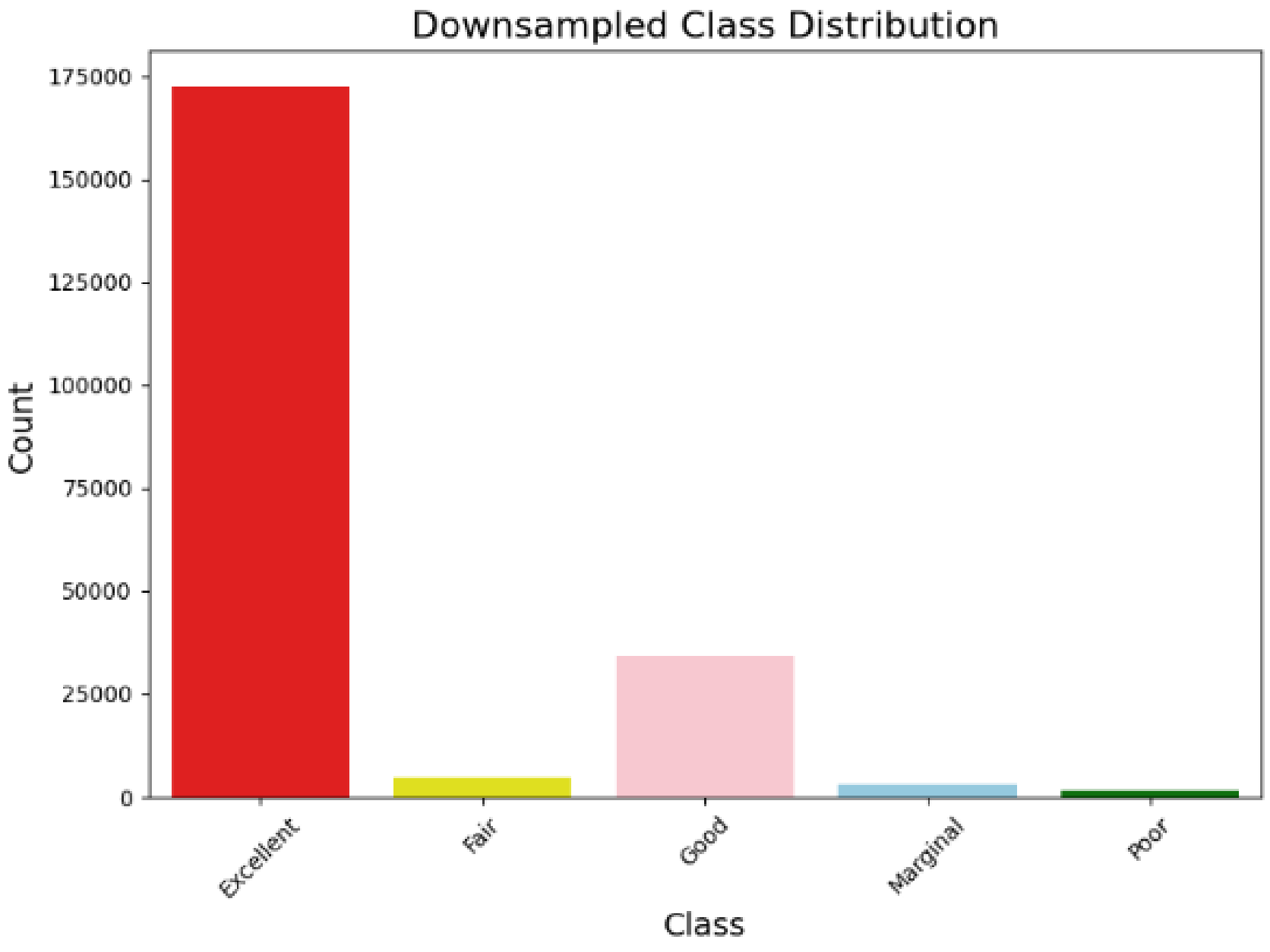

**Fig 3:** Classes down-sampled distribution

A useful technique for handling data imbalance is the undersampling. By reducing the 70% of volume on the world's combined dataset near balance for each classes were made. The models were rigorously tested for the entire dataset and reduction was applied on dataset. This also ensured the large data distribution for majority class wouldn't cause harm to performance and learning for minority classes on model.

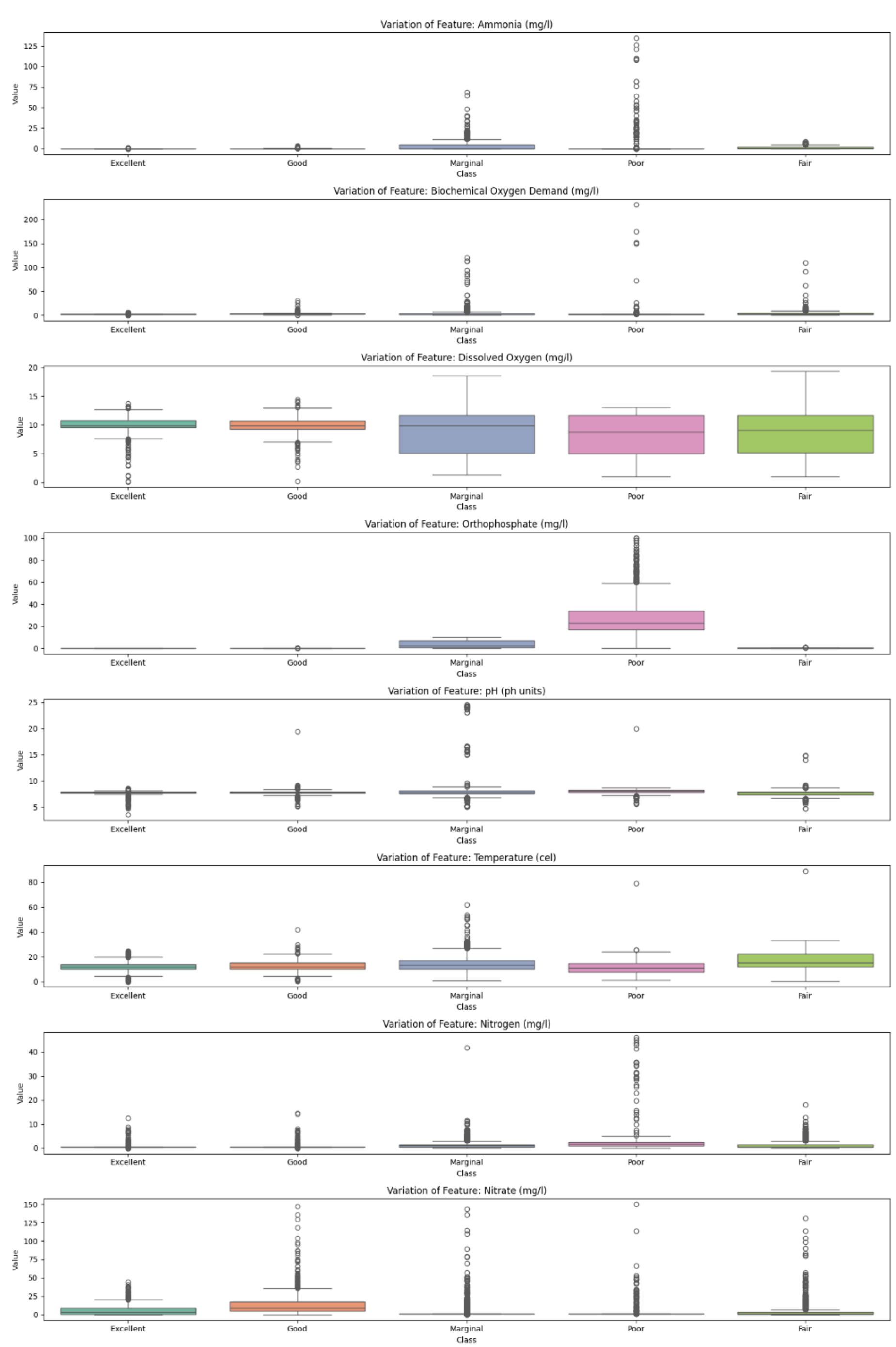

Fig 4 (a). First 1000 samples plot for Canada dataset

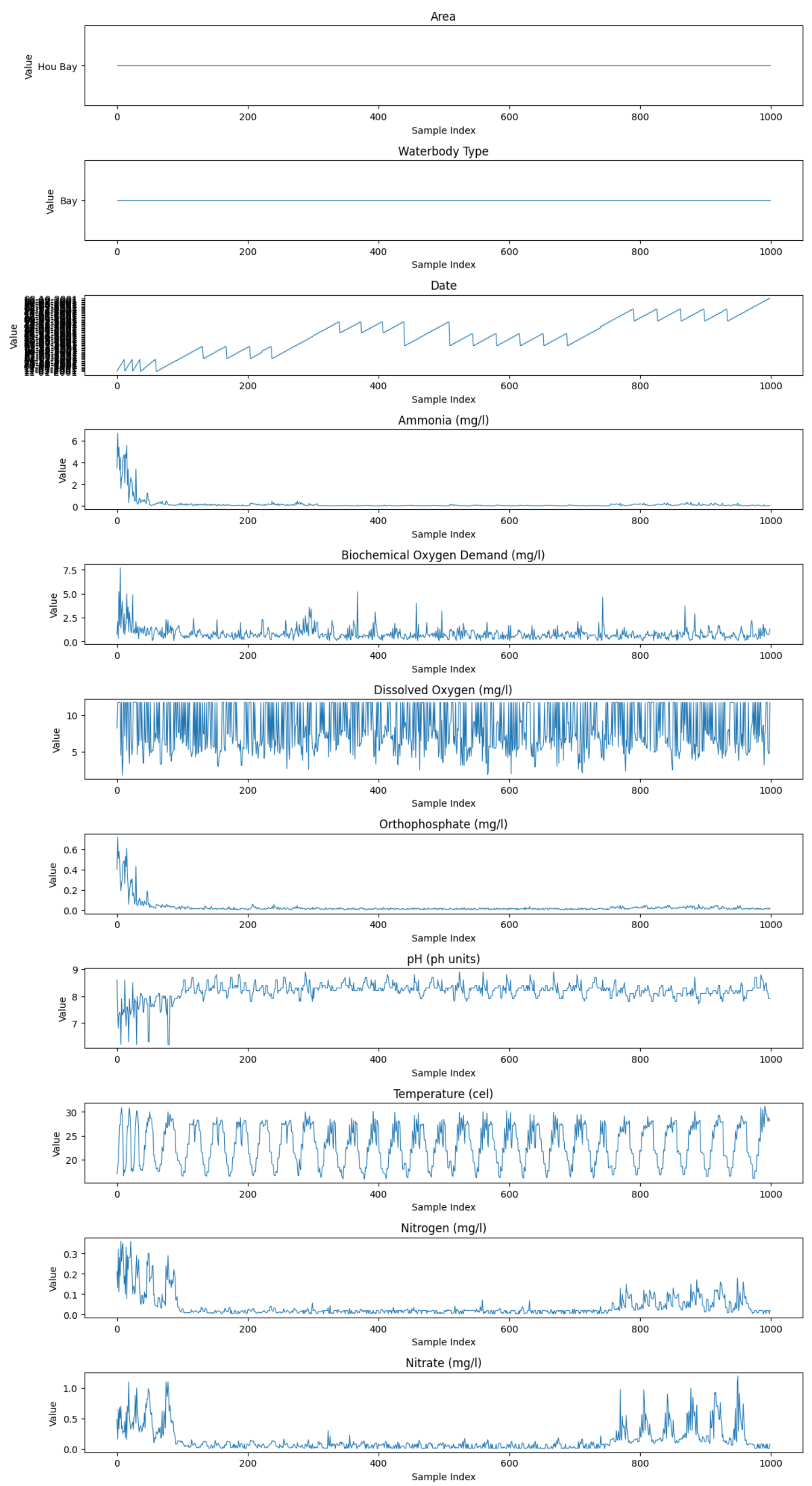

Fig 4 (b): Parametric variation in China's dataset (first 1000 samples)

Range (IQR) method. In this process, outliers for each feature were identified and removed independently based on the statistical threshold defined as values below

$Q1-1.5 \times IQR$ (1)

or above $Q3+1.5 \times IQR$ (2)

where Q1 and Q3 represent the first and third quartiles, respectively. Additionally, the first and last columns of the dataset were excluded to focus solely on the core numerical features. The cleaned plot reveals a more consistent and interpretable pattern across parameters. For instance, turbidity and pH now exhibit values within expected environmental ranges, with extreme spikes eliminated. Parameters like electrical conductivity, total alkalinity, and total hardness display structured variations, indicating differences between water sources while excluding previously observed anomalies. Interestingly, total iron, total ammonia, and chloride show minimal or no visible variation, likely due to the removal of extreme values, leaving behind uniform or nearly uniform measurements. Free residual chlorine (F.R.C.) retains some variability, though its range appears more controlled. Overall, the outlier removal process has enhanced the quality and interpretability of the dataset, making it more suitable for subsequent analysis and modeling.

Further, label encoder was used to handle the class encoding representations for performing the Machine learning based classification. As, data was classified into 3 classes as given in [10] we do not needed to apply any specific formulae for purpose of WQI calculation or evaluation.

A Water Quality Index (WQI) was computed and a map was created based on various physical water quality indicators measured in the study area. An index offers a composite influence of specific water quality measures on the overall quality of water in an area, making it a useful way to distill enormous amounts of data to their most basic form [12]. The table 2. Shows the WQI values classification phenomenon applied. The classes distribution was made 3 classes for katmandu city with Poor, Very Poor and Unsuitable for Drinking classes were combined into one category due to presence of mostly excellent water quality in reports published in KUKL (Kathmandu Upatyaka Khanepani Limited reports) [32].

Table 1: Water Quality classes distribution conditions

| Category Name | WQI Range (Inclusive) | Condition Logic | Description / Meaning |
|---|---|---|---|
| **Insufficient Data** | NaN | wqi is NaN (missing or invalid) | Not enough data to classify water quality |
| **Excellent** | 0 ≤ WQI ≤ 25 | wqi <= 25 | Very good water quality, safe for drinking |
| **Good** | 25 < WQI ≤ 50 | wqi <= 50 | Good quality, minor concerns |
| **Poor** | 50 < WQI ≤ 75 | wqi <= 75 | Some pollution, caution advised |
| **Very Poor** | 75 < WQI ≤ 100 | wqi <= 100 | Water quality degrading, not recommended for drinking |
| **Unsuitable for Drinking** | 100 < WQI ≤ 150 | wqi <= 150 | Contaminated, unsuitable for consumption |
| **Extremely Contaminated** | WQI > 150 | wqi > 150 | Highly polluted, dangerous to health |

**Standard Formula Used for Water Quality Index (WQI) Calculation**

The core formula for computing the Quality Index (Qi) for each parameter:

$$Q_i = \frac{(V - V_i)}{(S_i - V_i)} \times 100 \qquad (1)$$

- **V** = Measured value of the parameter in the sample.
- $S_i$ = Standard permissible value for the parameter (upper limit).
- $V_i$ = Ideal value for the parameter (usually zero or the ideal level).

Then, the overall **WQI** is calculated as a weighted average of all the Qi values:

$$WQI = \frac{\Sigma(Q_i \times W_i)}{\Sigma(W_i)} \qquad (2)$$

- $W_i$= Weight assigned to each parameter reflecting its importance.

**Labelling technique:**

1. Data Cleaning:
   The dataset is loaded using python dataframe pandas, and the water quality parameters are converted to numeric, with non-numeric and negative values handled appropriately.
2. Parameter-Level Quality Index (Qi) Calculation:
   For each water parameter in a sample, the code calculates a quality index $Q_i$ using the normalized difference from the ideal and standard values.
3. Handling Exceedances:
   If measured values exceed permissible limits significantly (more than twice the standard), additional penalties are applied to reflect the severity.
4. Weighted Aggregation to Compute WQI:
   The individual Qi values are weighted by their importance (Wi), summed, and normalized by the total weights to compute an overall WQI score per sample.
5. Classification of Water Quality:
   Based on the computed WQI, water samples are classified into categories from "Excellent" to "Extremely Contaminated," reflecting increasing pollution levels.
6. Critical Flags and Risk Assessment:
   Additional checks flag critical conditions (e.g., bacterial contamination, severe iron or turbidity levels). The risk level (Minimal to High Risk) is determined by combining WQI values and critical flags.
7. Summary and Reporting:
   The script developed in python outputs statistics, category distributions, risk levels, and saves enhanced results for further analysis.

By combining several metrics with domain-relevant standards, this method offers a methodical and quantitative technique to evaluate water quality and makes it simple to translate complex water quality data into actionable risk categories.

### 2.3 Hybrid Deep Learning Model

The proposed model integrates CNN and LSTM layers to capture both spatial and temporal dependencies in the water quality data.

• RNN Layers: The RNN component, a Long Short-Term Memory (LSTM) network with CNN, processes the sequential input to model temporal dependencies. This makes it easier to spot seasonal trends and variations in the water quality.

• Fully Connected Layers: The output from the CNN and LSTM layers is combined and processed by fully connected layers to provide the final predictions.

### 2.4 Model evaluation metrics:

**Accuracy:** It is the percentage of accurate findings—true positives and true negatives—among all instances that were looked at.

$\frac{TN+TP}{(FP+FN+TP+TN)}$...........Equation 1.

**Recall*:*** $\frac{TP}{(FP+FN)}$...........Equation 2.

The proportion of all real positive observations to all accurately projected positive observations

**Precision:** $\frac{TP}{(TP+FP)}$………….*Equation 3.*

The proportion of accurately forecast positive observations to the total number of positive predictions

**F1-score:**

$2*\frac{Precision*Recall}{Precision+Recall}$………….*Equation 4.*

It is the precision and recall weighted average, which balances the two

Confusion Matrix: It is a table that displays the counts of true positives, true negatives, false positives, and false negatives that is used to describe how well a classification model performs.

**Ten-fold Stratified Cross Validation:** A technique for assessing a model in which the dataset is divided into ten equal folds, each of which keeps the same percentage of classes, and the model is tested and trained ten times.

Nested Cross-Validation: This technique allows you to adjust a machine learning model's hyperparameters as you assess the model's performance. Cross-validation is used in two layers: an outer loop for model evaluation and an inner loop for hyperparameter adjustment. The data is separated into training and testing sets by the outer loop, and the optimal hyperparameters are determined by further splitting the training set by the inner loop. This method guarantees that hyperparameter tuning is carried out independently of the final evaluation, which helps minimize overfitting and offers a more accurate estimation of the model's performance.

**RMSE (Root Mean Squared Error):**

$RMSE$ =The prediction error magnitude of the model is indicated by the square root of the average of squared discrepancies between the actual and predicted values, or RMSE (Root Mean Squared Error).

$$RMSE = \sqrt{\frac{1}{n}\sum_{a=1}^{n}(x_a - y_a)}$$ ………………………….Equation 5.

Here, , $x_a$, $y_a$, and $x_m$ refer to the observed, measured, and average values for each n observation respectively.

**R² squared error:**

In a regression model, the percentage of the dependent variable's variance that can be predicted from the independent variables is called $R^2$ (or R-squared).

$$R^2 = \frac{\sum_{a=1}^{n}(x_a - x_m)^2 - \sum_{a=1}^{n}(x_a - y_a)^2}{\sum_{a=1}^{n}(x_a - x_m)^2}$$ …………………..Equation 6.

## 2.5 Model Training and Evaluation

The model was trained using a dataset split into training and testing sets, ensuring an equal representation of all seasons. Hyperparameters, such as batch size, learning rate, and dropout rate, were optimized using grid search. The model's performance was evaluated using metrics such as Root Mean Square Error (RMSE), and R-squared ($R^2$) score. The overall workflow is given in figure 3.

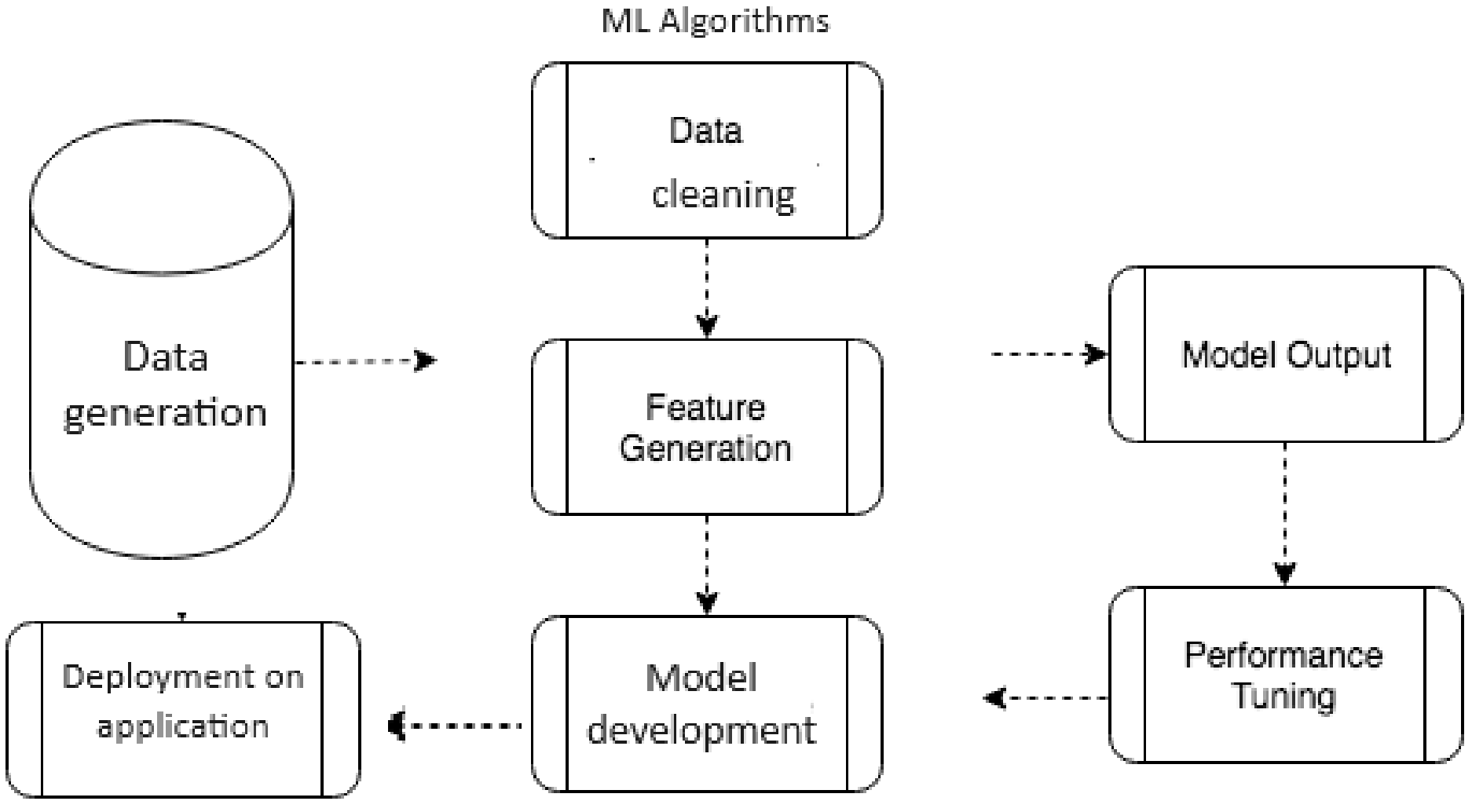

Fig 5: Proposed workflow

**For classification problem:**

Using the Keras library and TensorFlow backend, this model's methodology entails creating and training a neural network for a regression challenge. The dataset utilized in this instance comes from the "wqidata500-final_resampled.csv" file, which documents water quality analysis. The data is first loaded and divided into the target variable (y) and features (X). StandardScaler is used to apply standardization to the features, ensuring that each input variable contributes equally to the training of the model.

The neural network design that was selected is a straightforward feedforward network that is used in Keras as a sequential model.

## 2.5 RAG architecture and implementation

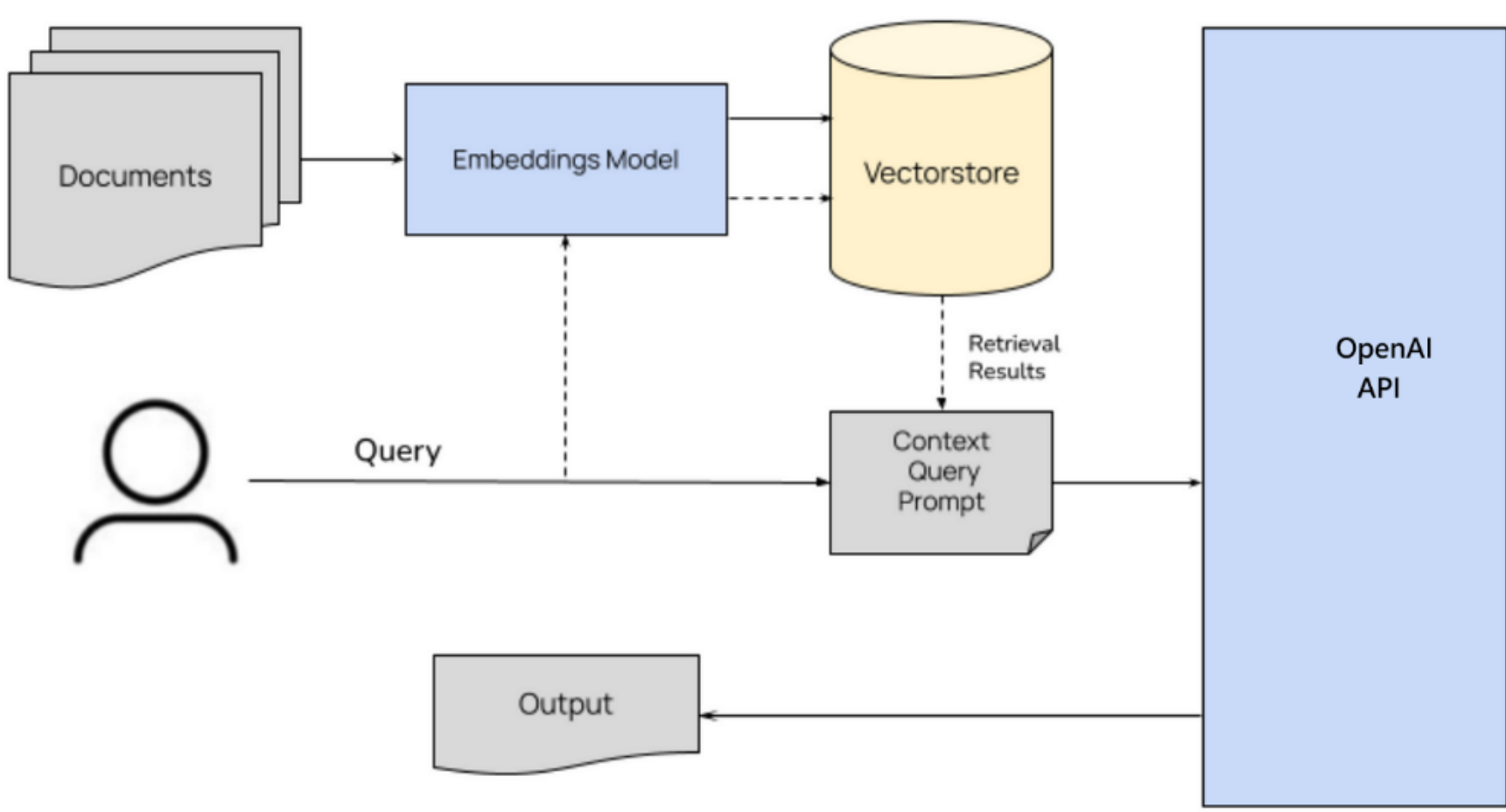

Fig 6: RAG architecture

The RAG (Retrieval-Augmented Generation) architecture in figure 6 is a novel approach that improves natural language comprehension and response generation by fusing the advantages of generative language models and retrieval-based systems. To give the model access to current facts and data, it first retrieves pertinent papers or information from a sizable external knowledge base. In order to provide contextually rich responses, this retrieval process is essential. The architecture's generative component successfully combines retrieved facts with language production skills to produce a logical and contextually applicable response after gathering the pertinent data.

A document pdf is utilized with relevant information on data and model related information as for predictions and results are then gained through OpenAI access through API token.

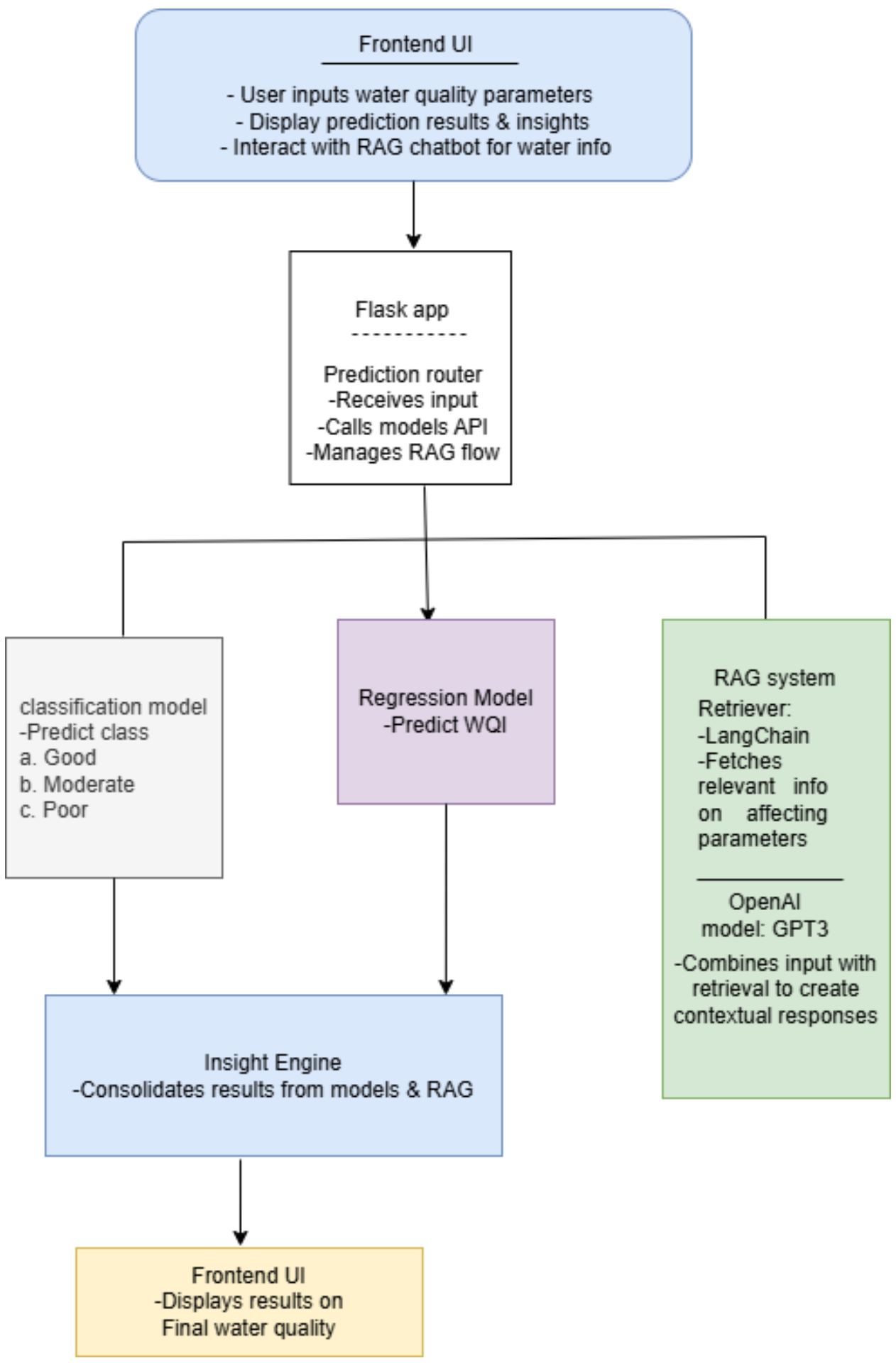

Fig 7: RAG chatbot development workflow

In addition to increasing the precision and applicability of the answers, this dual strategy allows the model to process a larger variety of questions, such as those requiring specialized expertise or current data. RAG is especially helpful in applications where customer pleasure depends on timely and pertinent information, such chatbots, question-answering systems, and interactive AI agents. RAG offers a more reliable and adaptable way to comprehend and reply to user inquiries in natural language by utilizing both retrieval and generation.

A Retrieval-Augmented Generation (RAG) architecture as seen in figure 7 can be implemented in a number of crucial processes in order to create a chat and information system pertaining to water quality. The system starts by pulling pertinent information from a PDF that contains data on water quality, including variables like pH, turbidity, temperature, and contaminant levels. After being processed and organized into a searchable way, this data is frequently kept in a vector database for quick and easy access. Based on user queries, relevant documents or sections are then found and retrieved from the knowledge base using the retrieval component of the RAG architecture. Concurrently, the generative model analyzes the information that has been retrieved to create logical, contextually appropriate answers to the user's questions on water quality concerns.

When a user inquires about the permissible pH values for safe drinking water, for example, the system pulls the relevant information from the organized database and provides a thorough response. This procedure promotes an engaging experience while guaranteeing that consumers receive timely and accurate information. Over time, the system's performance is improved by the ongoing feedback loop from user interactions, which further improves the retrieval and generation procedures.

## 3. Results and Discussions

In this section various models performances are compared and SHAPLEY Explanations results are presented. The chatbot as well as web application is developed and demonstrated for simulating the predictions.

### 3.1 Model performances

**Gaussian Naïve Bayes:**

The model Gaussian NB obtained a train Accuracy of 97.97% and a testing accuracy of 98.01%. The confusion matrix plot is shown in figure 15 below.

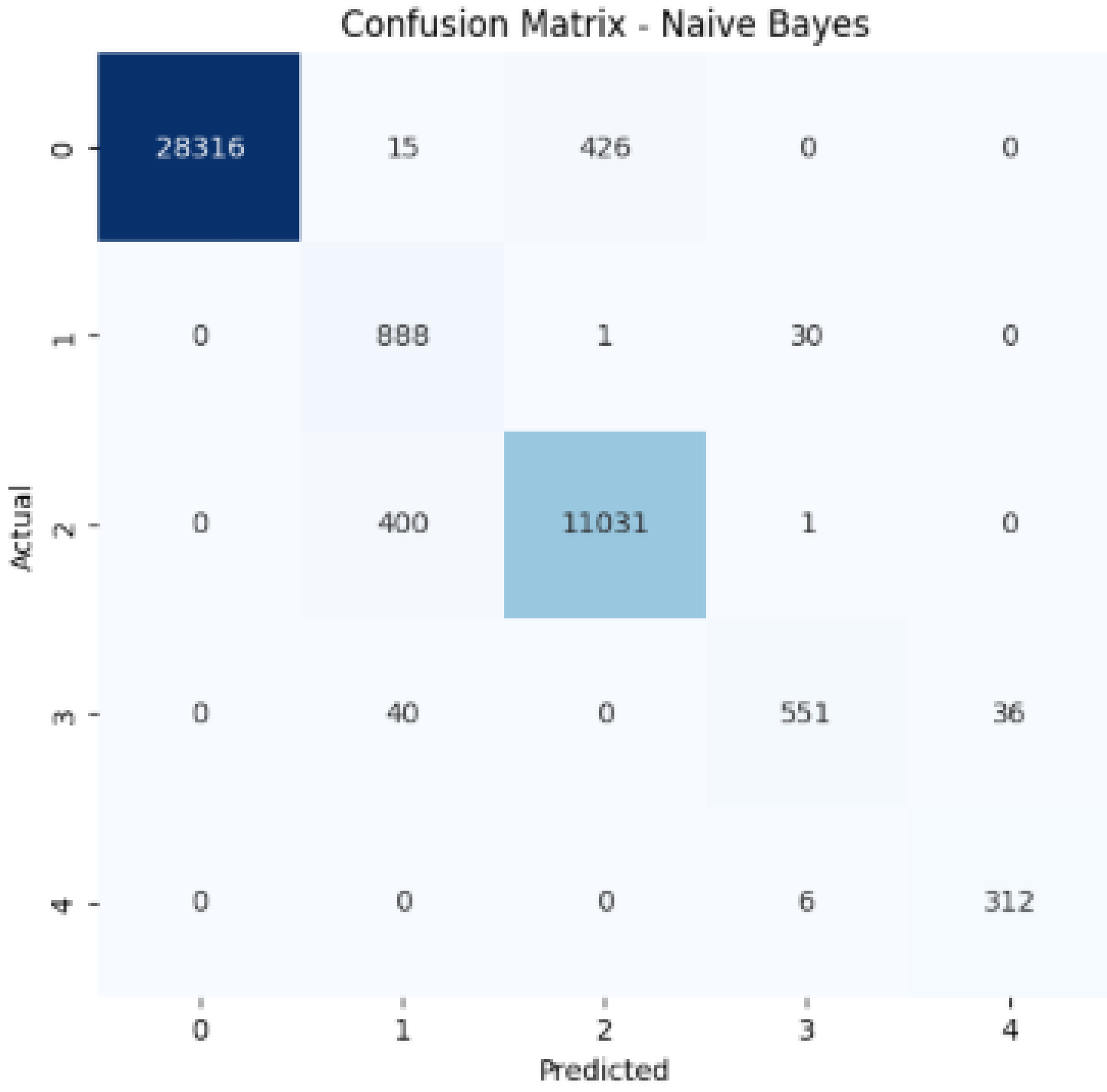

Fig 15: Confusion matrix plot

Table 3: Model hyperparameters for GNB model

| Hyperparameter | Description | Default Value |
|---|---|---|
| priors | Prior probabilities of the classes. If not specified, learned from data. | None |
| var_smoothing | Portion of the largest variance of all features added to variances for stability. Helps with numerical issues. | 1e-9 |

**Table 4:** Classification report for Gaussian Naïve Bayes model

| Class | Precision | Recall | F1-Score | Support |
|---|---|---|---|---|
| 0 | 1.00 | 0.99 | 0.99 | 34,509 |
| 1 | 0.75 | 0.96 | 0.84 | 919 |
| 2 | 0.93 | 0.96 | 0.95 | 6,859 |

| 3 | 0.92 | 0.88 | 0.90 | 627 |
|---|---|---|---|---|
| 4 | 0.90 | 0.98 | 0.94 | 318 |
| **Accuracy** | | | **0.98** | 43,232 |
| **Macro avg** | 0.90 | 0.95 | 0.92 | 43,232 |
| **Weighted avg** | 0.98 | 0.98 | 0.98 | 43,232 |

The models cross validation accuracies and performance on each fold is given in table as: This result shows that mean accuracy of approximately 98% can be obtained and standard deviation of only 0.0007 shows a great outcome and shows that data is less prone to outliers. This consistency indicates model stability and that the model is not heavily influenced by small changes in the data (like outliers). The nested cross validation result for model is given in table 5.

Table 5: Nested cross validation for model

| Fold | Test Accuracy |
|---|---|
| Fold 1 | 0.9814 |
| Fold 2 | 0.9799 |
| Fold 3 | 0.9796 |
| Fold 4 | 0.9799 |
| Fold 5 | 0.9787 |
| Fold 6 | 0.9802 |
| Fold 7 | 0.9798 |
| Fold 8 | 0.9793 |
| Fold 9 | 0.9796 |
| Fold 10 | 0.9803 |
| **Mean** | **0.9799** |
| **Std Dev** | **0.0007** |

**XGBoost classifier:**

The model XGboost obtained a training accuracy of 99.30% and test Accuracy of 98.99%. The model confusion matrix plot is given in figure 16. The plot shows that 34441 instances are correctly predicted for class '0' similarly, 806 instances are predicted correctly for class '1' and 6690 instances are predicted correctly for class '2', finally, 554 instances correctly predicted for class '3' and ' 303' instances are correctly predicted for class 4. The table 6 shows classification report for model.

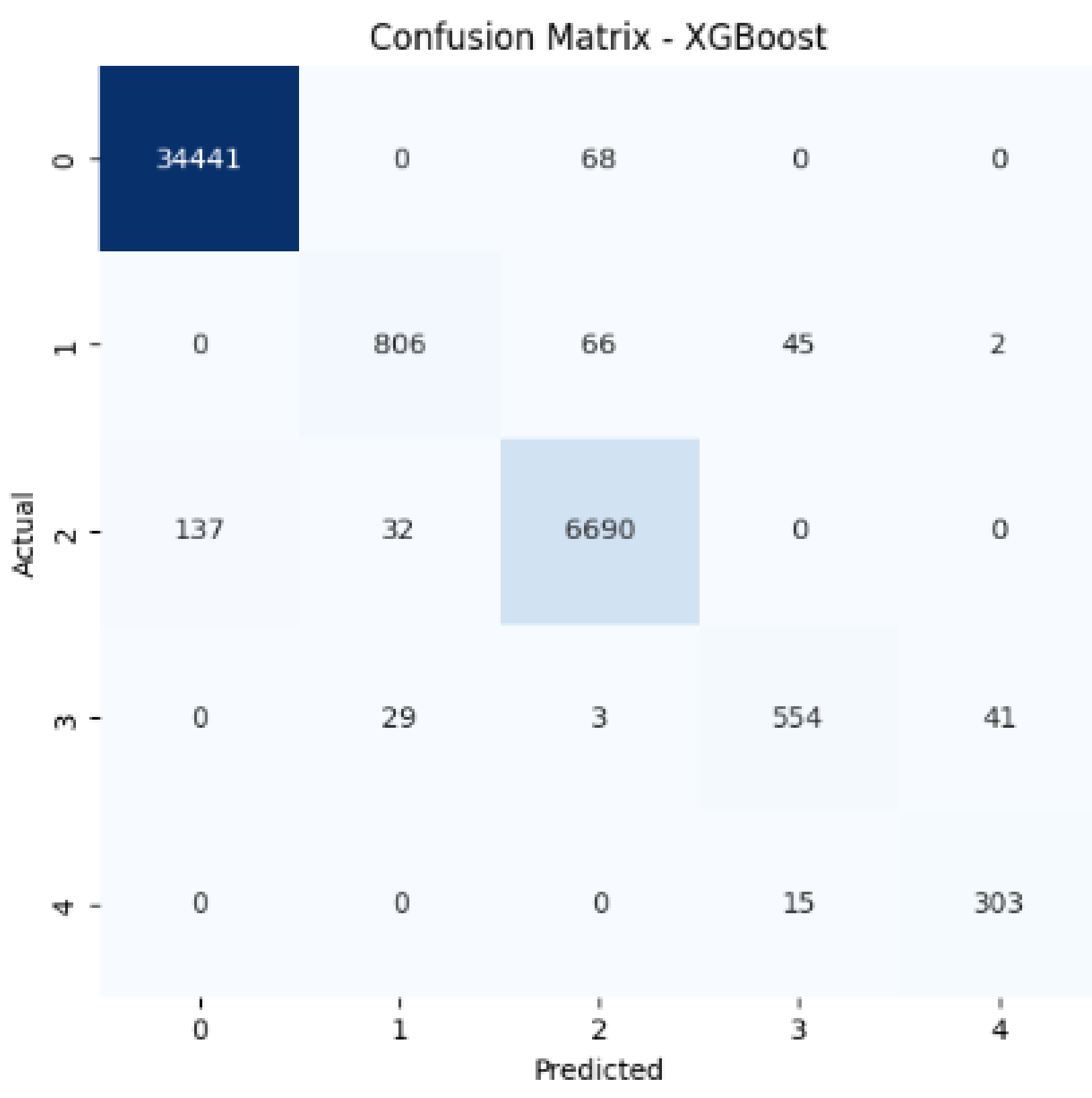

Fig 16: Confusion matrix plot

Table 6: Classification report for model

| Class | Precision | Recall | F1-Score | Support |
|---|---|---|---|---|
| **0** | 1.00 | 1.00 | 1.00 | 34,509 |

| 1 | 0.93 | 0.88 | 0.90 | 919 |
|---|---|---|---|---|
| **2** | 0.98 | 0.98 | 0.98 | 6,859 |
| **3** | 0.90 | 0.88 | 0.89 | 627 |
| **4** | 0.88 | 0.95 | 0.91 | 318 |
| **Accuracy** | | | **0.99** | 43,232 |
| **Macro Avg** | 0.94 | 0.94 | 0.94 | 43,232 |
| **Weighted Avg** | 0.99 | 0.99 | 0.99 | 43,232 |

**Bi-GRU with attention:**

The model Bi-GRU with attention demonstrated an excellent accuracy of 99.15% on testing set. This result shows that model generalizes well when downsampled and is good to recognize the new datasets.

**Table 7:** Tabulated version of Bi-GRU with attention model classification report

| Class | Precision | Recall | F1-Score | Support |
|---|---|---|---|---|
| Excellent | 0.99 | 1.00 | 1.00 | 34,509 |
| Fair | 0.92 | 0.97 | 0.94 | 919 |
| Good | 0.99 | 0.96 | 0.98 | 6,859 |
| Marginal | 0.98 | 0.95 | 0.96 | 627 |
| Poor | 0.99 | 0.99 | 0.99 | 318 |
| **Accuracy** | | | **0.99** | 43,232 |
| **Macro Avg** | 0.98 | 0.97 | 0.97 | 43,232 |
| **Weighted Avg** | 0.99 | 0.99 | 0.99 | 43,232 |

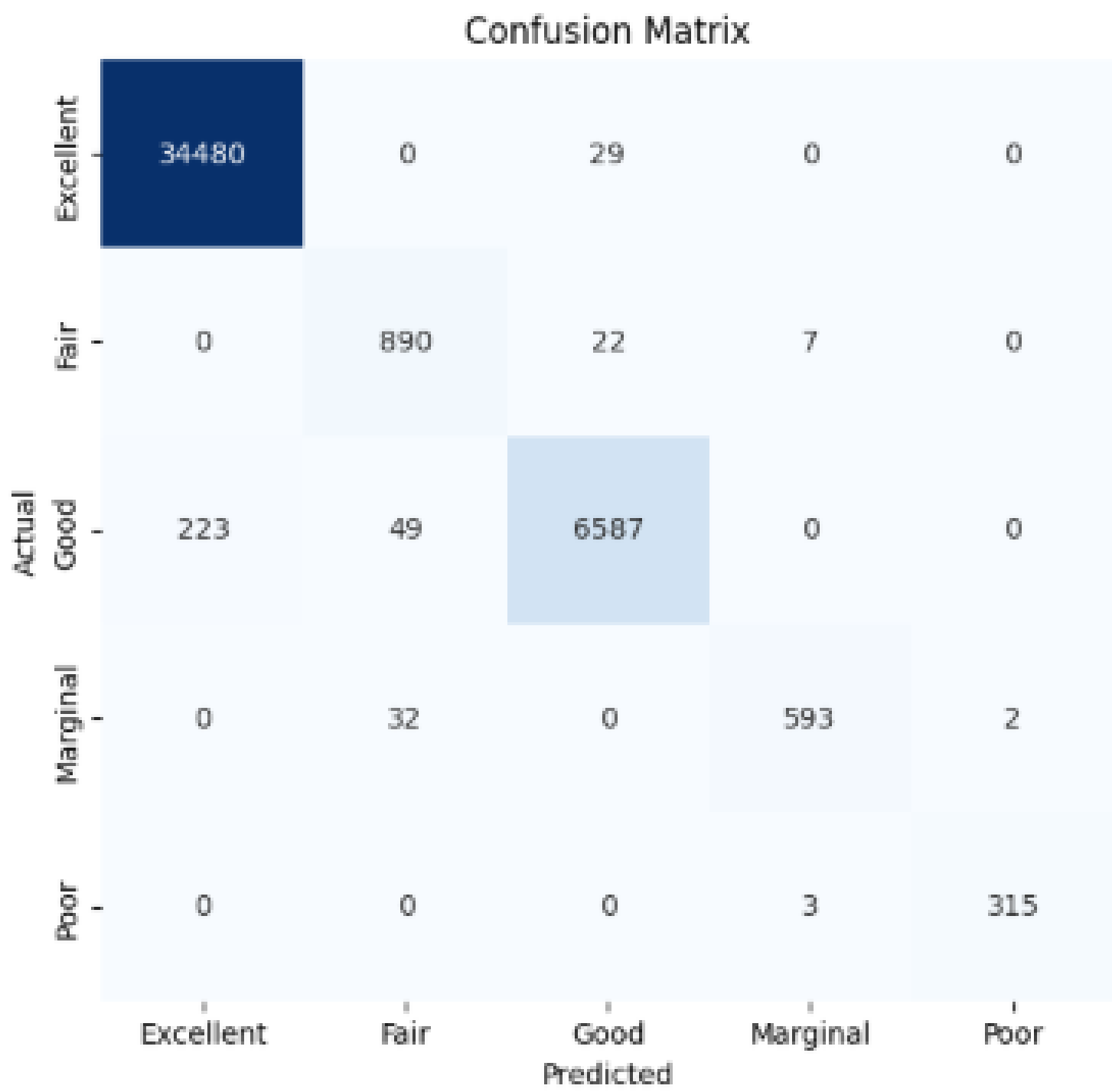

Fig 17: Confusion matrix plot

This confusion matrix in figure 17 visually summarizes the performance of a classification model across five categories: Excellent, Fair, Good, Marginal, and Poor. The diagonal values indicate the number of correct predictions for each class, showing the model's strong ability to identify "Excellent" (34,480) and "Good" (6,587) instances, as well as decent performance for "Fair" (890), "Marginal" (593), and "Poor" (315). Off-diagonal entries highlight misclassifications; notably, 223 "Good" instances were incorrectly labeled as "Excellent" and 29 "Excellent" instances were misclassified as "Good," suggesting some confusion between these two categories. The relatively low numbers of misclassifications for "Poor" and "Excellent" indicate the model excels at identifying these extreme ends of the spectrum, while more errors occur between the middle categories.

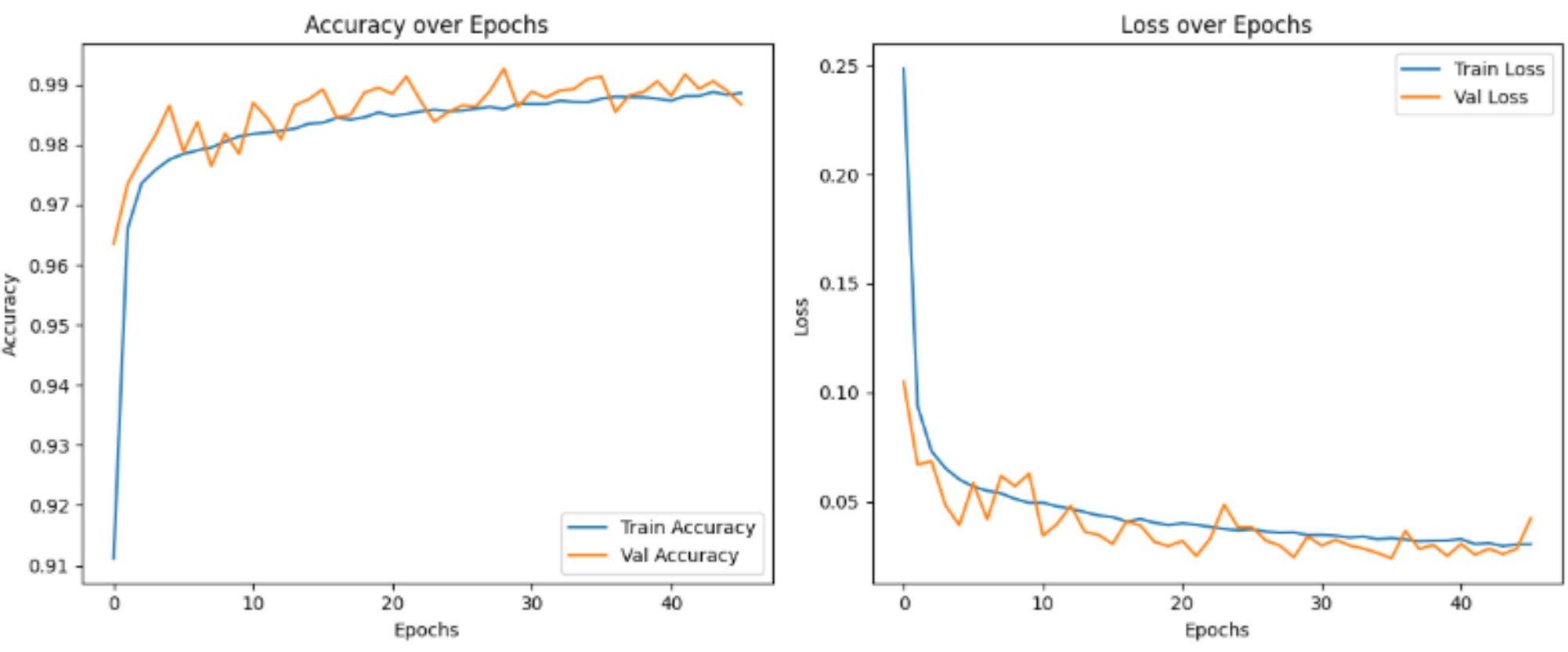

Fig 18: Training and loss history plot

The figure 18 shown training history plots illustrate the performance of a Bi-GRU attention model over 45 epochs by tracking both accuracy and loss on the training and validation datasets. In the left plot (Accuracy over Epochs), the model's training accuracy shows a steady upward trend, starting around 91% and reaching over 98.5%. The validation accuracy also increases rapidly in the early epochs and remains consistently high, slightly fluctuating between 98% and 99%. The close alignment of training and validation accuracy indicates that the model is learning well and generalizing effectively, with no major signs of overfitting.

In the given right plot (Loss over Epochs), both the training loss and validation loss decline significantly in the initial epochs, showing that the model is minimizing errors during training. After about 10 epochs, the loss curves flatten out and continue to decline more gradually, with some minor fluctuations in validation loss. However, validation loss remains lower than or similar to training loss throughout, which further suggests that the model is stable and generalizes well without overfitting or underfitting.

**Hybrid Convolutional Neural Network with Long Short-Term Memory (CNN-LSTM) for Multiclass Water Quality Classification:**

The model after choosing 30% of entire data for class 0 and class ‘2’ obtained 99.20% accuracy demonstrating superior performance to Bi-directional GRU with attention model. This shows that CNN can be useful for identifying water quality data and create pattern for superior output performance.

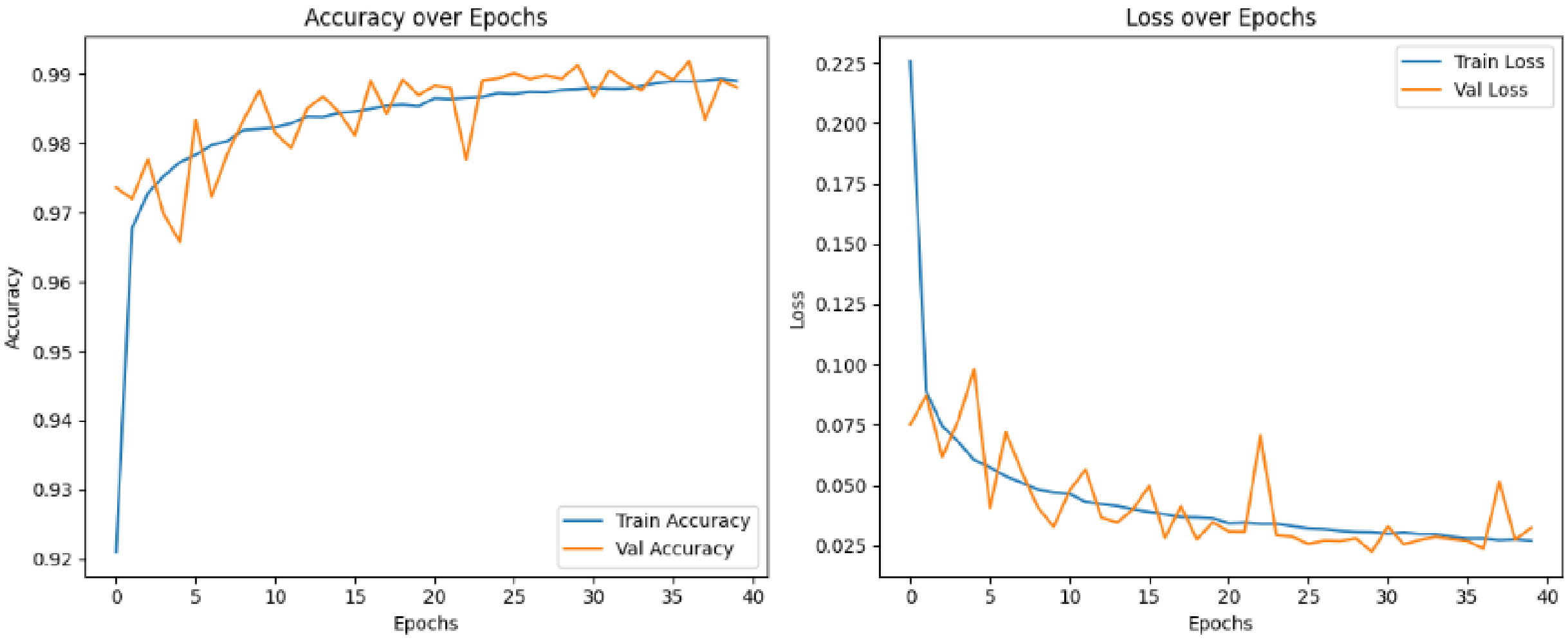

Fig 19: CNN-LSTM model accuracy and loss history plot.

The provided training and loss history plots in figure 19 illustrate the performance of a CNN-LSTM over 39 epochs by tracking both accuracy and loss on the training and validation datasets. In the left plot (Accuracy over Epochs), the model's training accuracy shows a steady upward trend, starting around 91% and reaching over 99%. The validation accuracy also increases rapidly in the early epochs and remains consistently high, slightly fluctuating between 97% and 99%. The curve pattern on validation accuracy and loss as shown in figure of training and validation accuracy indicates that the model is learning well and generalizing effectively, with no major signs of overfitting.

In the given right plot (Loss over Epochs)**,** both the training **loss** and validation **loss** decline significantly in the initial epochs, showing that the model is minimizing errors during training. After about 10 epochs, the loss curves flatten out and continue to decline more gradually, with some minor fluctuations in validation loss. However, validation loss remains lower than or similar to training loss throughout, which further suggests that the model is stable and generalizes well without overfitting or underfitting.

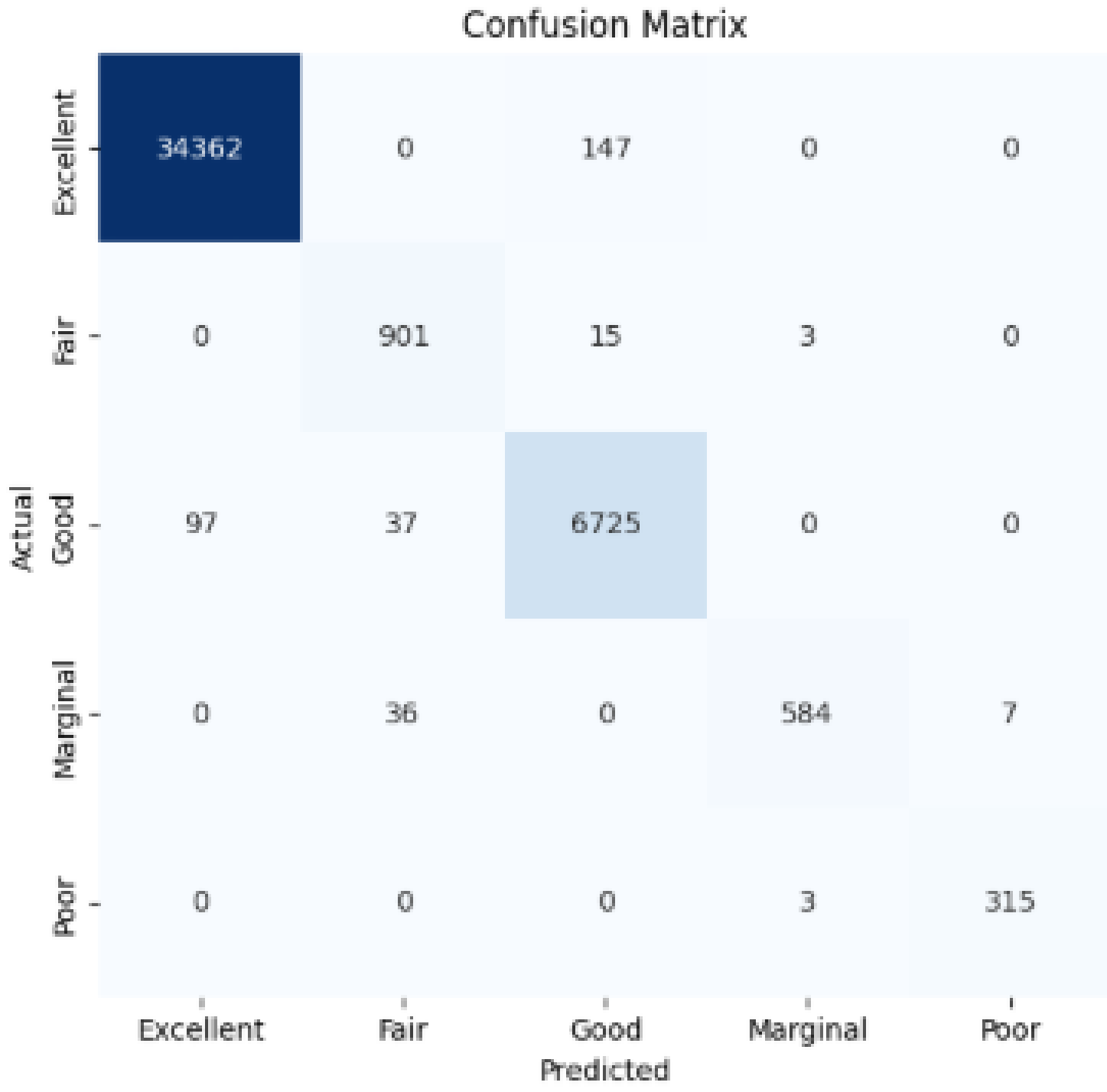

Fig 20: Confusion matrix plot

This confusion matrix in figure 20 visually summarizes the performance of a classification model across five categories: Excellent, Fair, Good, Marginal, and Poor. The diagonal values indicate the number of correct predictions for each class, showing the model's strong ability to identify "Excellent" (34,362) and "Good" (6725) instances, as well as decent performance for "Fair" (901), "Marginal" (584), and "Poor" (315). The relatively low numbers of misclassifications for "Poor" and "Excellent" indicate the model excels at identifying these extreme ends of the spectrum, while more errors occur between the middle categories.

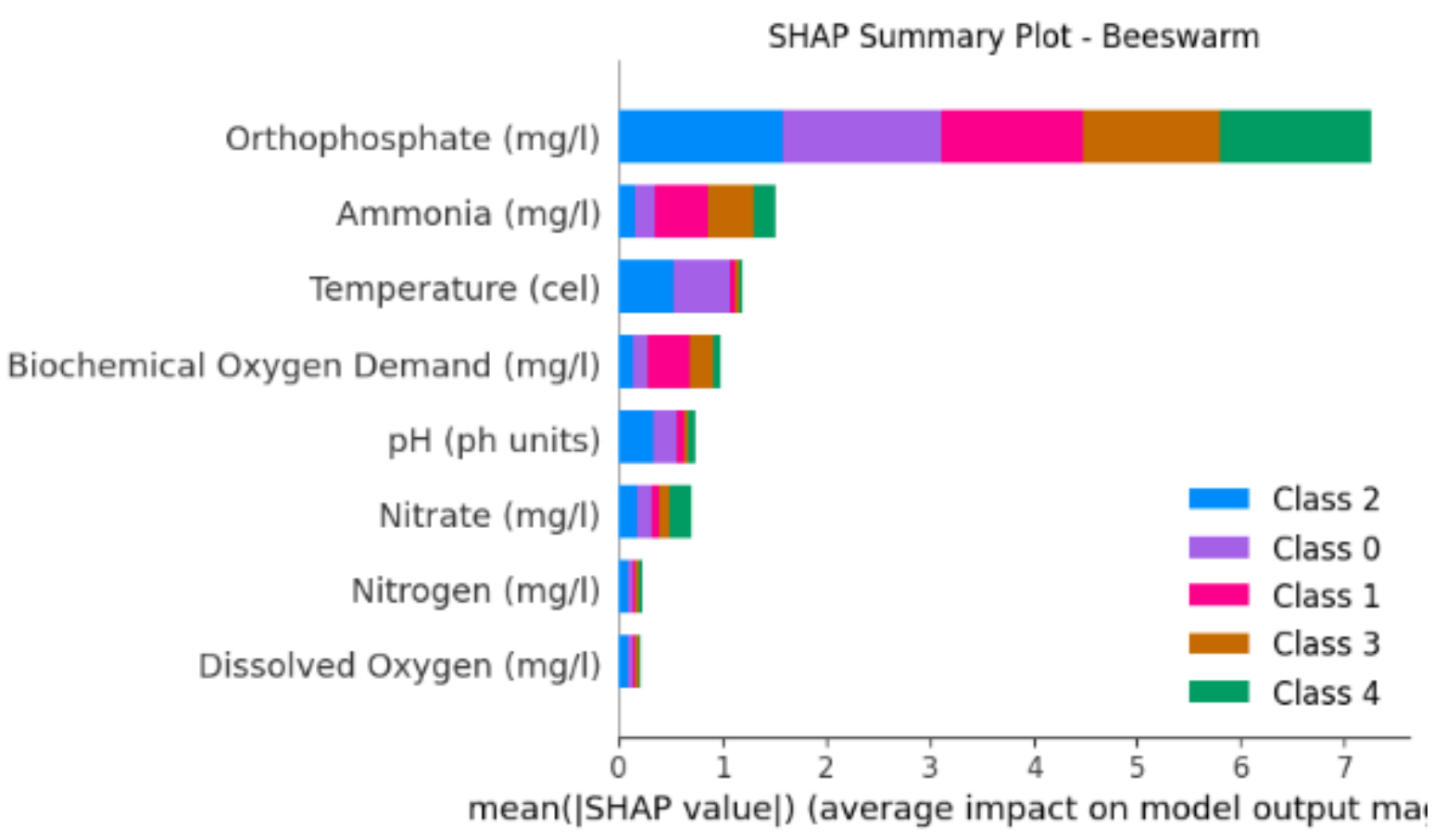

**Fig 21 (a):** SHAP XAI analysis plot

This SHAP summary beeswarm plot in figure 21 (a) visualizes the impact of each water quality parameter on the output of a multi-class classification model used to predict water quality classes. Each bar on the y-axis represents a different input feature (e.g., Orthophosphate, Ammonia, Temperature), and the x-axis shows the mean absolute SHAP value, which quantifies the average contribution of each feature to the model's prediction. The longer the bar, the more influence that feature has on the classification outcome.

From the plot, Orthophosphate (mg/l) emerges as the most impactful feature by a significant margin, playing a major role across all classes (Class 0 to Class 4), particularly Class 4. This indicates that the level of orthophosphate is a critical determinant in classifying water quality. Following it, Ammonia and Temperature are also influential, though to a lesser extent. Features like Biochemical Oxygen Demand (BOD), pH, and Nitrate contribute moderately, while Nitrogen and Dissolved Oxygen show the least impact on the model's decisions. The color-coded sections within each bar denote how much each class was influenced by that feature, providing insight into class-specific feature importance. Thus, this SHAP plot in figure 21 (a) and 21 (b) helps in interpreting which parameters are mostly significantly affecting for the classification decisions made by the model and can guide water quality management and policy decisions.

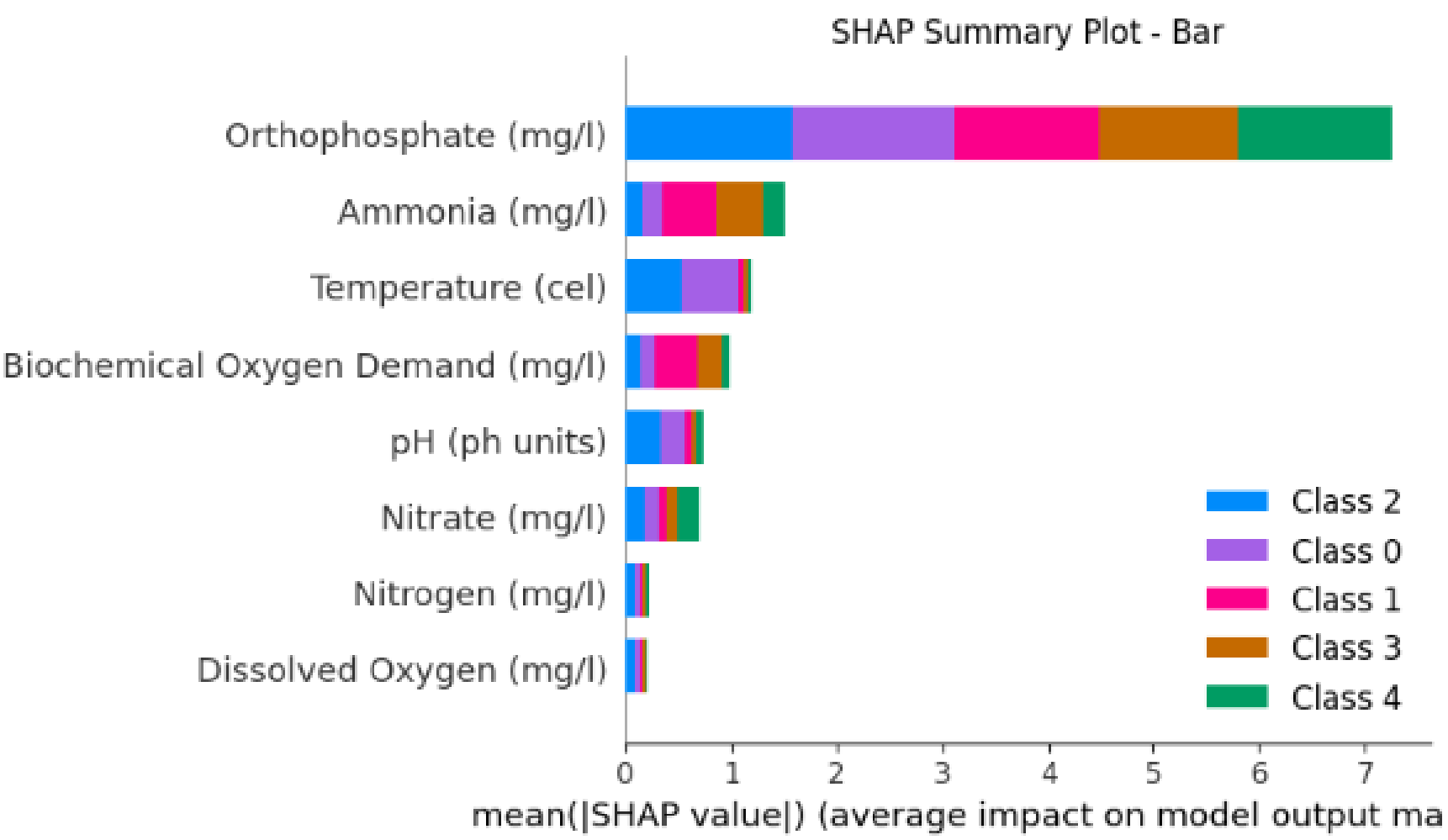

**Fig 21 (b):** SHAP magnitude bar plot

**Table 8:** Hyperparameters wise values

| Model | Hyperparameter | Value |
|---|---|---|
| CNN-LSTM | Conv1D Filters | 64 |
| | Kernel Size | 1 |
| | MaxPooling1D Pool Size | 1 |
| | LSTM Units | 128 |
| | Dense Units | 64 |
| | Dropout Rate | 0.3 |
| | Activation Functions | ReLU (hidden), Softmax (output) |
| | Optimizer | Adam |
| | Loss Function | Categorical Crossentropy |
| | Batch Size | 32 |
| | Epochs | 200 |
| | EarlyStopping Patience | 10 |
| | Validation Split | 0.2 |
| Bi-GRU + Attention | GRU Units | 128 |

| | Bidirectional | Yes |
|---|---|---|
| | Attention Layer | Custom Layer (Softmax-based) |
| | Dense Units | 64 |
| | Dropout Rate | 0.3 |
| | Activation Functions | ReLU (hidden), Softmax (output) |
| | Optimizer | Adam |
| | Loss Function | Categorical Crossentropy |
| | Batch Size | 32 |
| | Epochs | 200 |
| | EarlyStopping Patience | 10 |
| | Validation Split | 0.2 |
| XGBoost | Number of Estimators | 100 |
| | Learning Rate | 0.01 |
| | Max Depth | 3 |
| | Colsample by Tree | 0.8 |
| | Subsample | 0.8 |

**Regression analysis:**

For regression analysis for predicting the values the labels classes were dropped in order to ensure that the dataset is complex problem. The Extra trees regressor obtained Test $R^2$ Score: 1.0000 and Test RMSE: 0.0021. Similarly, the catboost regressor obtained a test $R^2$ Score: 0.9999 and test RMSE: 0.0426.

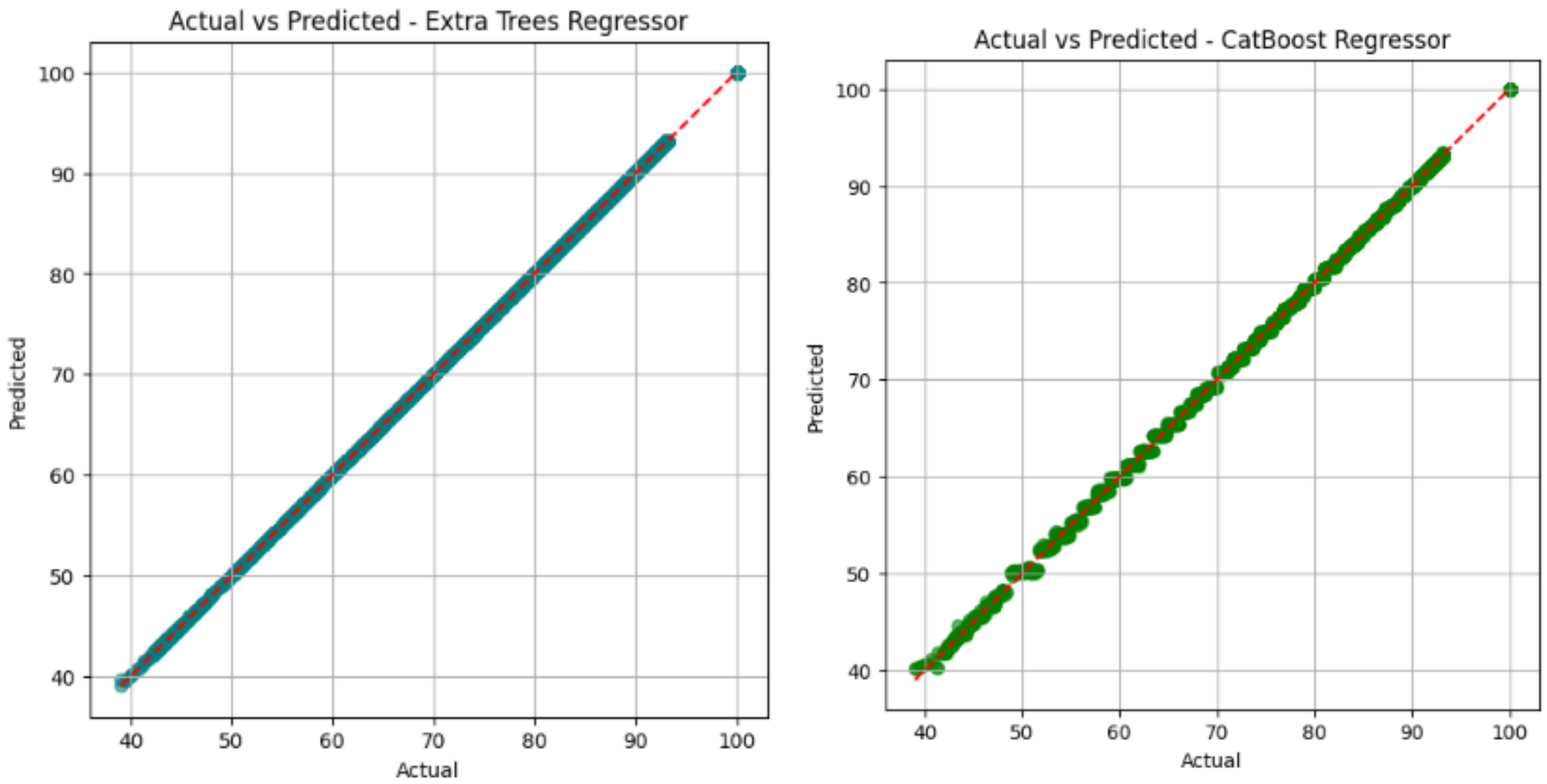

Fig 22 (a): Actual vs predicted plot for Extra Trees regressor    Fig 22 (b): Actual vs predicted pl
ot for catboost regressor

This result as shown in table 9 shows the potential for real world prediction of CCME value with greater performance.

Table 9: Machine Learning Field for Water Quality Prediction:

| Use Case | Significance |
|---|---|
| **Label Generation** | CCME-WQI is commonly used to classify water samples (e.g., into Good/Poor) — useful for supervised classification models**.** |
| **Target Variable** | Can be used as a regression target for predicting WQI from raw sensor or environmental data. |
| **Data Integration** | Integrates multiple water quality features into one metric, which helps when using single-output models**.** |
| **Model Explainability** | SHAP, LIME, and other tools can show how each parameter **(e.g.,** nitrate, pH) affects WQI, improving interpretability. |
| **Benchmarking** | CCME-WQI acts as a standard benchmark for comparing ML models across datasets. |

| **Policy-Driven AI** | Models trained to predict WQI can support policy-making, early warning systems, and resource prioritization. |
|---|---|

## 3.3 Prediction application

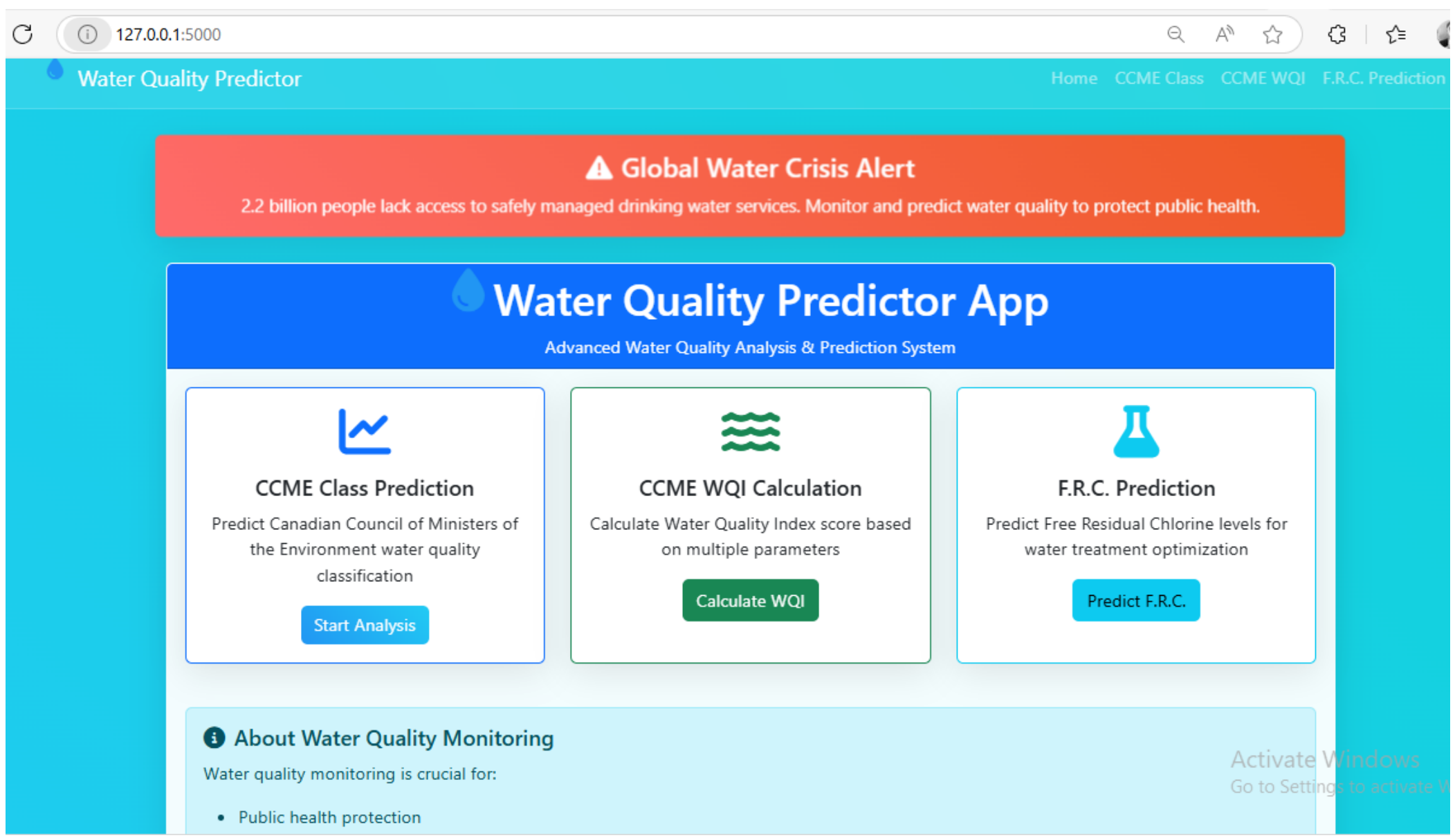

Fig 23(a). Multi-functional application UI for classification

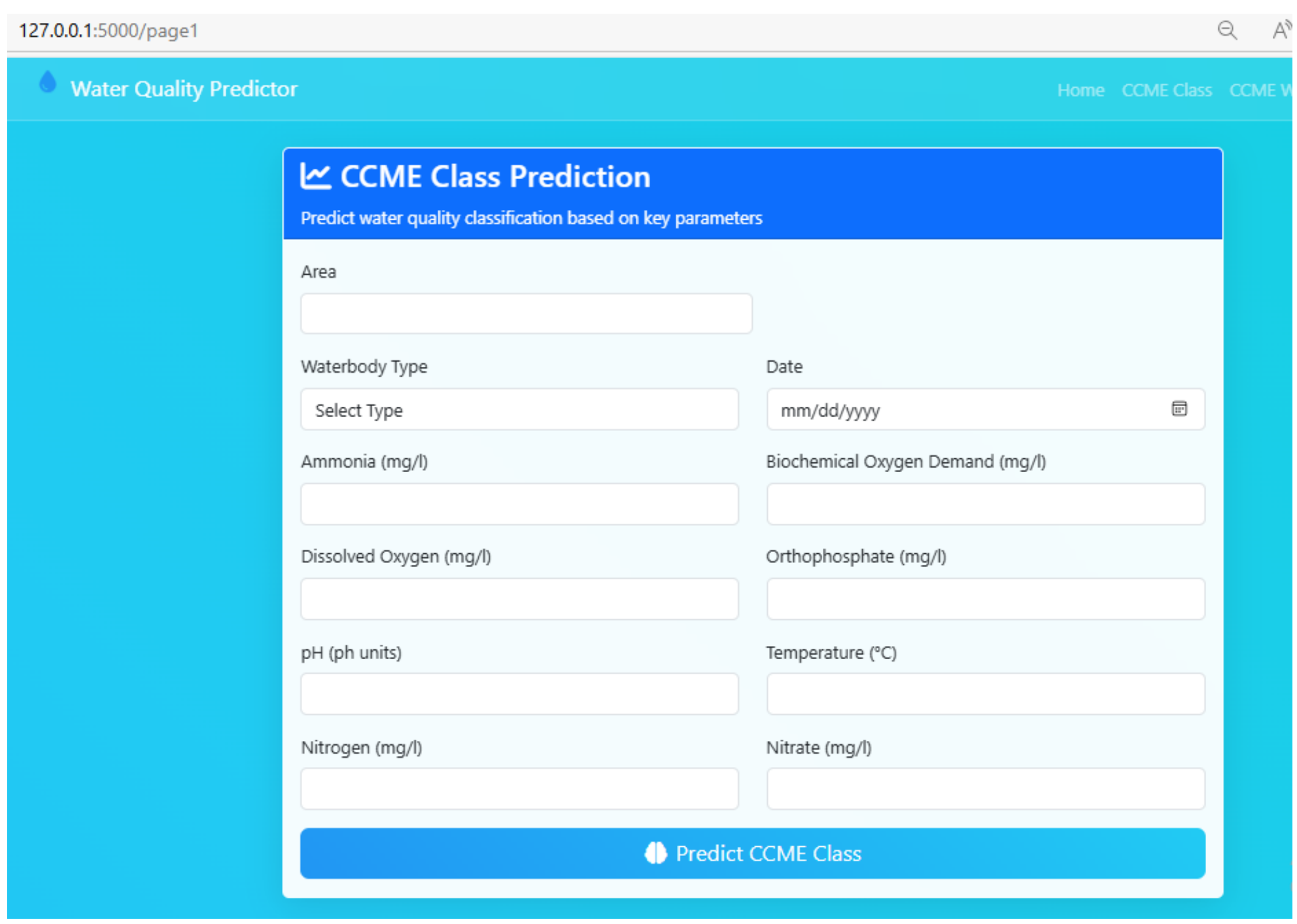

Fig 23(b) WQI app for regression task

The Flask-powered Water Quality Predictor App shown in figure 23 (a) and figure 23(b) offers an intuitive user interface for forecasting and categorizing Water Quality Index (WQI) values according to predefined criteria, and the UI in Figures. 20(a) & 20(b) shows the input fields for predictions. "WQI Prediction" and "WQI Classification" are the two primary pages of the app. Each page has a gradient sky blue navigation bar with links to "Home," "WQI Classification," "WQI Prediction," and "Exit," set against a gradient black background for a polished look. Two trained models (classification and regression) are implemented through the backend for prediction and classification.

### i. WQI Classification Page:

Users can enter values for four water quality metrics on this page. The app processes the inputs (placeholder logic for actual classification) upon pressing the sky blue "Predict" button. Based on established thresholds, the app then presents the predicted classification (e.g., "Good", "Average", "Poor"). With the use of the given characteristics, this page attempts to classify water quality into easily understood categories.

### ii. WQI Prediction Page:

Users can enter values for the same four parameters on this page. The application processes the inputs (placeholder logic for actual regression) and displays the projected WQI value when the "Predict" button is clicked. This figure shows the result of calculating the composite influence of the input parameters on the overall quality of the water using a regression model (placeholder logic). Based on the data supplied, this page offers a numerical evaluation of the water quality.

## 3.4 RAG implementation

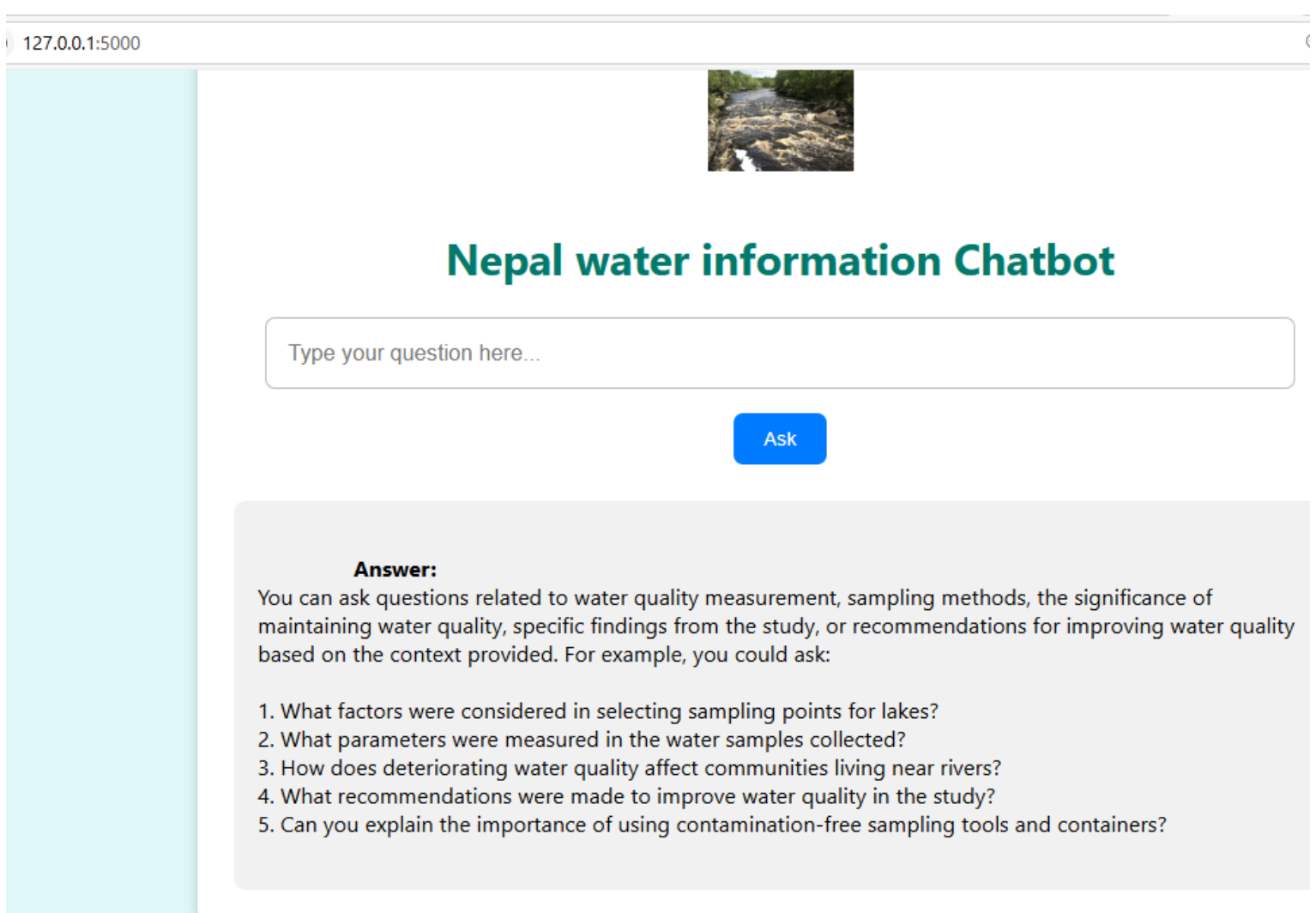

Figure 24: Chatbot conversation

The chatbot as seen in figure 24 is able to apply the predictions for providing insights on water quality and exhibiting measures for the purpose of water quality and hygiene related information because of the various model integrations with RAG Chabot. Overall, the Water Quality Predictor App offers a smooth experience to users who want to learn about water quality classifications and receive predictive insights into WQI values by seamlessly integrating Flask for backend operations. This chatbot is a Retrieval-Augmented Generation (RAG)-based PDF question-answering system developed using Flask, LangChain, FAISS, and OpenAI's GPT models. The application follows a typical RAG pipeline by combining document retrieval with generative language modeling to provide accurate, context-aware responses. It begins by loading environment variables OpenAI API key and specifying the path to the PDF file (water quality pdf report 2077 obtained from [26]. The chatbot extracts raw text from the PDF using the PyMuPDF (fitz) library and splits the content into semantically meaningful chunks using LangChain's Recursive Character Text Splitter, which ensures that each chunk retains contextual relevance through overlapping.

These text chunks are then converted into dense vector embeddings using OpenAI's embedding model. The resulting vectors are stored in a FAISS vector store, which serves as the retriever component of the RAG architecture. When a user submits a question via the Flask-based web interface, the system retrieves the most relevant chunks from the FAISS index based on semantic similarity to the query. These retrieved chunks are then passed to the generator—the OpenAI GPT-4o-mini model—via LangChain's RetrievalQA chain. The model uses both the query and the retrieved context to generate a grounded, natural language response. The final answer is displayed on the webpage, enabling interactive, document-specific conversations. By leveraging the RAG paradigm, the chatbot effectively combines knowledge retrieval with language generation to enhance the reliability and accuracy of its responses. The usage of RAG is an insightful approach for environmental monitoring and water management decision-making because of its easy-to-use interface and simple navigation.

The results show different ML models effectively capture the complex spatial and temporal relationships in water quality data. The integration of MLP and RNN layers allows for comprehensive feature extraction and sequence modeling, enhancing the prediction accuracy. This approach can be extended to other regions and water quality datasets, providing a robust framework for environmental monitoring. The LIME explainable AI [18, 20] provides insights on various parameters for decision-making and support for the users. The usage of the flask application for making predictions and gaining insights via a RAG implementation is a useful concept for gaining hints on various approaches for enhancing water sustainability.

## 4. Limitations and future scopes

The assessment using the global CCME dataset can be vital for Machine learning modeling and classification of water-related parameters. The use of hybrid architectures can be insightful for the development of advanced applications, as shown in this work. With 99.20% accuracy hybrid architectures such as CNN-LSTM can be vital in AI-based water quality assessment. The RAG-based chatbot can provide users of this application with detailed insights and ultimately benefit from predicting water quality for informed use. Nonetheless, to fully comprehend water quality for more extensive community health and environmental management initiatives, thorough investigations integrating chemical and microbiological analyses are advised. ML algorithms can be very useful for classifying the water quality of a specific area and predicting WQI using the given parameters [5-12]. In the future, the work can be expanded to a broader dataset with a wide range of features, and models can be designed and developed for a vast array of data, including IoT (Internet of Things) and edge-based systems for monitoring water quality and performing predictive analysis. Future research can explore the incorporation of additional environmental factors and sensors for more data collection. Methods for various factors, such as rainfall and land-use patterns, are employed to further enhance prediction accuracy. Moreover, real-time monitoring and prediction systems can be developed to provide timely alerts and support decision-making processes.

Utilizing IoT and edge-based systems and increasing sensors for more detailed parametric analysis would be a key scope of this project which can be enhanced in future [2]. The ongoing water dryness is also one of a great burden causing massive water scarcity in the global world, the heatwave and many such environmental related issues can be fought by appropriately applying measures for maintaining water quality [5]. Using more advanced AI models such as, explainable AI models [15], developing a product level app usable by the general public would be also the priority after the completion of this research work. The future research could cover entire world-based research on WQI parameters.

## 5. Conclusion

This research presents an explainable AI-based analysis and predictions through machine learning for water quality, and the models utilized showcased improved accuracy and reliability for CCME dataset from major developed nations. The results highlight the potential of machine learning in environmental applications, paving the way for better water resource management and policy development. Further, the RAG chatbot provides insights and details on water quality pollutants and remedies as well as risk factors etc.

The water quality being very important parameter for enhancing a country landscape the prediction yielding a good result through machine learning models including interpretability analysis with SHAP suggest a great feasibility in solving global water pollution related crisis by effectively mitigating the issue.

**Authors Declarations section:**

**Author's contributions statement**

- Biplov Paneru: Methodology, conceptualization, data collection, software, Data curation, validation, write original draft
- Bishwash Paneru: validation, testing and validating, software, Data curation, write original draft, final draft preparation

The authors would like to declare that they have equally contributed for the research study and are fully aware for publishing of this research study.

**Disclosure of interest**

The authors declare that there are no known competing interests

**Declaration of funding statement**

Authors declare that they received no financial aid or funding for this work

**Data availability statement**

The data would be made available on reasonable request

**Clinical Trial Number:** Not applicable.

**Ethics, Consent to Participate, and Consent to Publish declarations:** Not applicable.

**Declaration of preprint availability**

The authors declare that the pre-print version of this work is available for users for early access at:

Arxiv: https://doi.org/10.48550/arXiv.2409.10898

(Authorea Preprints) Techrxiv: 10.36227/techrxiv.172684041.14494773/v1

WaterQualityNeT: Prediction of Seasonal Water Quality of Nepal Using Hybrid Deep Learning Models - TechRxiv

**References**

[1] Bashyal, P., Adhikari Pradhananga, M., Adhikari, N., 2023. Time and space domain prediction of water quality parameters of Bagmati River using deep learning methods. *BIBECHANA*, 20(3), 248-258. https://doi.org/10.3126/bibechana.v20i3.57736

[2] Kuroki, S., Ogata, R., Sakamoto, M., 2023. Predicting the presence of E. coli in tap water using machine learning in Nepal. *Water and Environment Journal*. https://doi.org/10.1111/wej.12844

[3] Abbas, F., Cai, Z., Shoaib, M., Iqbal, J., Ismail, M., Arifullah, Alrefaei, A.F., Albeshr, M.F., 2024. Machine learning models for water quality prediction: A comprehensive analysis and uncertainty assessment in Mirpurkhas, Sindh, Pakistan. *Water*, 16(7), 941. https://doi.org/10.3390/w16070941

[4] Alqahtani, A., Shah, M.I., Aldrees, A., Javed, M.F., 2022. Comparative assessment of individual and ensemble machine learning models for efficient analysis of river water quality. *Sustainability*, 14(3), 1183. https://doi.org/10.3390/su14031183

[5] S., Y., A., M., 2021. Efficient water quality prediction for Indian rivers using machine learning. *Asian Journal of Applied Science and Technology*, 5(11), 100-109. https://doi.org/10.38177/ajast.2021.5111

[6] Shams, M., Elshewey, A., El-kenawy, E.S., Ibrahim, A., Talaat, F.M., Tarek, Z., 2023. Water quality prediction using machine learning models based on grid search method. *Multimedia Tools and Applications*, 83. https://doi.org/10.1007/s11042-023-16737-4

[7] Sangwan, V., Bhardwaj, R., 2024. Application of machine learning model for assessing water quality index. In: Yadav, A.K., Yadav, K., Singh, V.P. (Eds.), *Integrated management of water resources in India: A computational approach*, vol. 129, pp. 325–348. Springer. https://doi.org/10.1007/978-3-031-62079-9_16

[8] Venkatesh, A.T., Rajkumar, S., Masilamani, U.S., 2024. Assessment of groundwater quality using water quality index, multivariate statistical analysis and machine learning techniques in the vicinity of an open dumping yard. *Environment, Development and Sustainability*. https://doi.org/10.1007/s10668-024-05209-w

[9] Bhattarai, R., Dahal, K., 2020. Review of water pollution with special focus on Nepal. *Journal of Nepal Geological Society*, 7, 101-111. https://doi.org/10.1729/Journal.23618

[10] Bohara, B., 2016. Water quality index of southern part of the Kathmandu Valley, Central Nepal; Evaluation of physical water quality parameters of shallow wells. *Bulletin of the Department of Geology*, 19, 45-56. https://doi.org/10.3126/bdg.v19i0.19989

[11] Ghimire, S., Pokhrel, N., Pant, S., Gyawali, T., Koirala, A., Mainali, B., Angove, M.J., Paudel, S.R., 2022. Assessment of technologies for water quality control of the Bagmati River in Kathmandu valley, Nepal. *Groundwater for Sustainable Development*, 18, 100770. https://doi.org/10.1016/j.gsd.2022.100770

[12] Gupta, I., Kumar, A., 2015. Detection and mapping of water quality variation in Godavari River using WQI, clustering and GIS. *Journal of Geographic Information System*, 7, 71-84.

[13] Khan, Y., See, C.S., 2016. Predicting and analyzing water quality using machine learning: A comprehensive model. *2016 IEEE Long Island Systems, Applications and Technology Conference (LISAT)*. https://doi.org/10.1109/LISAT.2016.7494106

[14] Ashfaq, F., Aman, U., 2024. Prediction of water quality using effective machine learning techniques. *Journal of Computers and Intelligent Systems*, 2(1). https://journals.iub.edu.pk/index.php/JCIS/article/view/2712

[15] Nallakaruppan, M.K., Gangadevi, E., Shri, M.L., et al., 2024. Reliable water quality prediction and parametric analysis using explainable AI models. *Scientific Reports*, 14, 7520. https://doi.org/10.1038/s41598-024-56775-y

[16] Iffaty, A., Salsabila, A., Aufa, A., Suhendra, R., Yusuf, M., Sasmita, N., 2023. Enhancing Water Quality Assessment in Indonesia Through Digital Image Processing and Machine Learning. *Grimsa Journal of Science Engineering and Technology*, 1, 1-8. https://doi.org/10.61975/gjset.v1i1.3

[17] Del Castillo, A.F., Garibay, M.V., Díaz-Vázquez, D., Yebra-Montes, C., Brown, L.E., Johnson, A., Garcia-Gonzalez, A., Gradilla-Hernández, M.S., 2024. Improving river water quality prediction with hybrid machine learning and temporal analysis. *Ecological Informatics*, 82, 102655. https://doi.org/10.1016/j.ecoinf.2024.102655

[18] Malek, N.H.A., Wan Yaacob, W.F., Md Nasir, S.A., Shaadan, N., 2022. Prediction of Water Quality Classification of the Kelantan River Basin, Malaysia, Using Machine Learning Techniques. *Water*, 14, 1067. https://doi.org/10.3390/w14071067

[19] Choden, Y., Chokden, S., Rabten, T., et al., 2022. Performance assessment of data driven water models using water quality parameters of Wangchu river, Bhutan. *SN Appl. Sci.*, 4, 290. https://doi.org/10.1007/s42452-022-05181-y

[20] Makumbura, R.K., Mampitiya, L., Rathnayake, N., Meddage, D., Henna, S., Dang, T.L., Hoshino, Y., Rathnayake, U., 2024. Advancing Water Quality Assessment and Prediction Using Machine Learning Models, Coupled with Explainable Artificial Intelligence (XAI) Techniques Like Shapley Additive Explanations (SHAP) For Interpreting the Black-Box Nature. *Results in Engineering*, 23, 102831. https://doi.org/10.1016/j.rineng.2024.102831

[21] Nasir, N., Kansal, A., Alshaltone, O., Barneih, F., Sameer, M., Shanableh, A., & Al-Shamma'a, A. (2022). Water quality classification using machine learning algorithms. *Journal of Water Process Engineering*, *48*, 102920. https://doi.org/10.1016/j.jwpe.2022.102920

[22] Rammohan, B.; Partheeban, P.; Ranganathan, R.; Balaraman, S. Groundwater Quality Prediction and Analysis Using Machine Learning Models and Geospatial Technology. *Sustainability* 2024, *16*, 9848. https://doi.org/10.3390/su16229848

[23] Garcia, J.; Heo, J.; Kim, C. Machine Learning Algorithms for Water Quality Management Using Total Dissolved Solids (TDS) Data Analysis. *Water* 2024, *16*, 2639. https://doi.org/10.3390/w16182639

[24] Bai, Y.; Xu, Z.; Lan, W.; Peng, X.; Deng, Y.; Chen, Z.; Xu, H.; Wang, Z.; Xu, H.; Chen, X.; et al. Predicting Coastal Water Quality with Machine Learning, a Case Study of Beibu Gulf, China. *Water* 2024, *16*, 2253. https://doi.org/10.3390/w16162253

[25] Madni, H.A.; Umer, M.; Ishaq, A.; Abuzinadah, N.; Saidani, O.; Alsubai, S.; Hamdi, M.; Ashraf, I. Water-Quality Prediction Based on $H_2O$ AutoML and Explainable AI Techniques. *Water* 2023, *15*, 475. https://doi.org/10.3390/w15030475

[26] Department of Hydrology and Meteorology. (2072 B.S. / published April 23, 2021). *Water Quality Study by DHM – 2072*. Government of Nepal, Department of Hydrology and Meteorology

[27] The Global Water Crisis is a Local Problem, UCL Risk and Disaster Reduction Blog. [Online]. Available: https://blogs.ucl.ac.uk/rdr/2023/03/22/the-global-water-crisis-is-a-local-problem/. [Accessed: Jul. 27, 2025].

[28] "36 per cent cities to face water crisis by 2050," *Down to Earth*, Sept. 2023. [Online]. Available: https://www.downtoearth.org.in/news/water/36-per-cent-cities-to-face-water-crisis-by-2050-91325. [Accessed: Jul. 27, 2025].

[29] F. Ghanem and Y. Mezran, *The Looming Climate and Water Crisis in the Middle East and North Africa*, Carnegie Endowment for International Peace, 2023. [Online]. Available: https://carnegieendowment.org/2023/06/07/looming-climate-and-water-crisis-in-middle-east-and-north-africa-pub-89977. [Accessed: Jul. 27, 2025].

[30] Zou, S.; Ju, H.; Zhang, J. Water Quality Management in the Age of AI: Applications, Challenges, and Prospects. *Water* 2025, *17*, 1641. https://doi.org/10.3390/w17111641

[31] Karim, M.R., Syeed, M.M.M., Rahman, A. *et al.* A Comprehensive Dataset of Surface Water Quality Spanning 1940-2023 for Empirical and ML Adopted Research. *Sci Data* **12**, 391 (2025). https://doi.org/10.1038/s41597-025-04715-4